\documentclass[11pt]{article}
\usepackage[margin=1in]{geometry}
\usepackage[hyphens]{url}
\usepackage{graphicx}
\usepackage{amsmath,amssymb}
\usepackage{booktabs}
\usepackage{placeins}
\usepackage{float}
\usepackage[numbers,sort&compress]{natbib}
\usepackage{authblk}
\usepackage{microtype}
\usepackage[hidelinks]{hyperref}
\title{muSync-GS: Physics-Synchronized Driving Video Synthesis for Weather and Geometric Road Hazards}
\author[1]{Yang Chen}
\author[1]{Yicheng Zhu}
\author[2]{Tao Li}
\author[1]{Zilin Bian\thanks{Corresponding author.}}
\affil[1]{Rochester Institute of Technology}
\affil[2]{City University of Hong Kong}
\date{August 2026}
\hypersetup{
  pdftitle={muSync-GS: Physics-Synchronized Driving Video Synthesis for Weather and Geometric Road Hazards},
  pdfauthor={Yang Chen, Yicheng Zhu, Tao Li, and Zilin Bian}
}

\begin{document}
\maketitle

\begin{abstract}

High-quality driving data are essential for autonomous-driving systems and generative world models. However, rare and safety-critical scenarios involving adverse weather, braking under low tire–road friction, and uneven road geometry are costly and risky to collect at scale. Existing video-generation and 3D Gaussian editing methods can modify weather appearance or road geometry, but typically do not couple these edits with tire--road interaction and vehicle dynamics. As a result, an edited video may retain its original trajectory even when the modified road condition should alter braking, wheel slip, load transfer, and ego-camera motion. We present muSync-GS, a physics-synchronized framework for driving video synthesis under adverse-weather and road-elevation hazards. A precipitation-derived road-surface condition jointly controls road appearance and tire friction, while a shared road-elevation profile drives both visible road-geometry editing and axle excitation. A calibrated vehicle model  predicts speed, slip ratio, normal loads, and pitch for constructing the ego-camera trajectory and synchronized physical annotations. On 12 held-out CarSim cases spanning precipitation levels, brake inputs, and
road-profile parameters, the model achieves mean case-wise RMSEs of
$0.0273\,\mathrm{m/s}$ for speed, $0.0590^\circ$ for pitch, $0.0101$ for slip ratio, and $26.61\,\mathrm{N}$ for per-wheel normal load. Together with the reconstructed-scene experiments, these results show that muSync-GS accurately reproduces vehicle responses under held-out controls while synchronizing them with controllable scene edits and
ego-camera motion.
\end{abstract}

\begin{figure}[t]
\centering
\includegraphics[width=0.85\columnwidth]{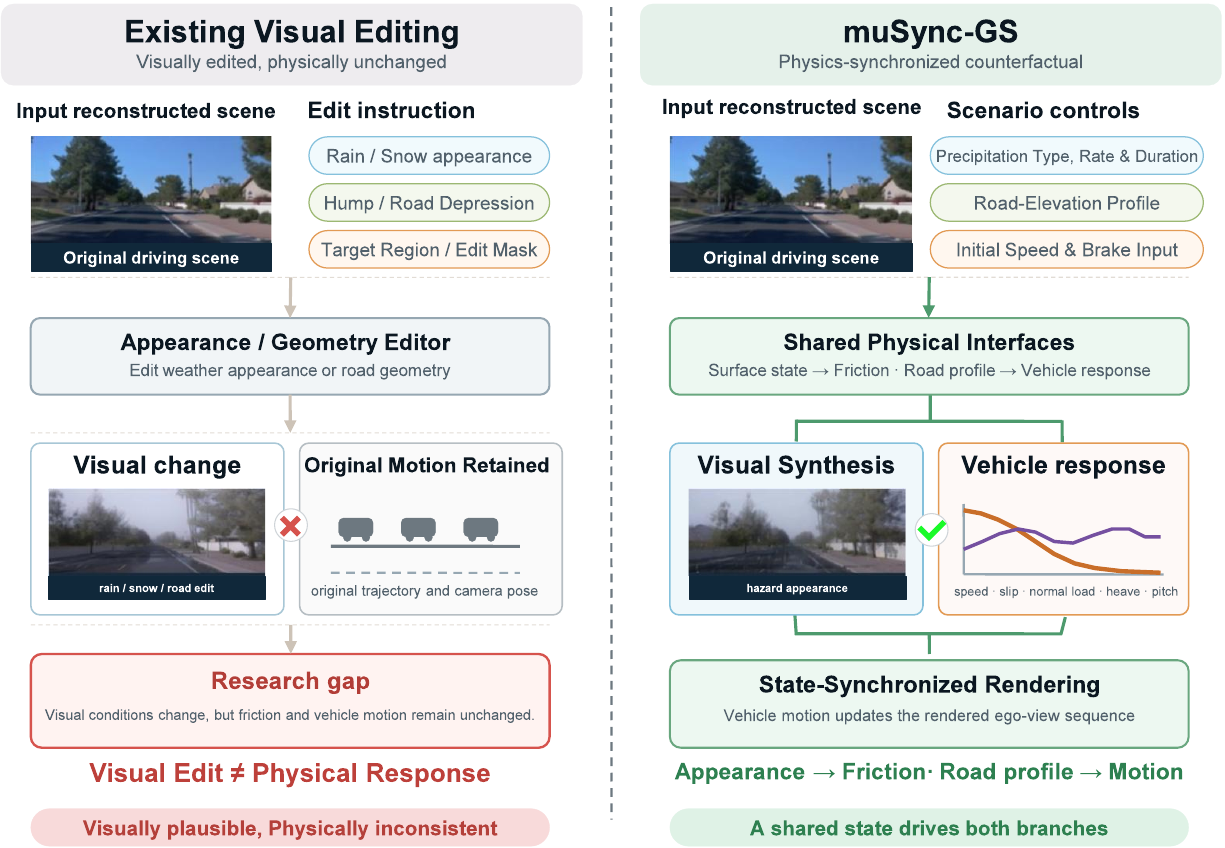}
\caption{Conceptual comparison between appearance-only editing and muSync-GS. Appearance-only editing modifies weather or road geometry while reusing the original ego-camera motion, whereas muSync-GS synchronizes appearance, tire--road interaction, vehicle response, and ego-camera motion through a shared scenario state.}
\label{fig:conceptual_gap}
\end{figure}


\section{Introduction}
Recent advances in autonomous driving have enabled data-driven driving policies and generative world models to perceive, predict, and simulate complex traffic environments~\cite{gaia1,panacea2024,drivedreamer2,dtdima2025,seeunsafe2025,ttcx2024,valik2025}. The development of these systems depends heavily on large-scale, diverse, and high-quality driving data, particularly data covering rare and safety-critical events. However, data from scenarios such as emergency braking on wet or snow-covered roads and driving over uneven road surfaces are costly and potentially dangerous to collect at scale. The associated physical states, including tire--road friction, slip ratio, normal loads, and vehicle attitude, are also difficult to obtain from ordinary driving datasets. These limitations motivate the need for controllable
synthetic data that capture not only how a driving scenario looks, but also
how the vehicle should physically respond.

Recent driving-video generation and editable 3D Gaussian methods can synthesize rain and snow at different severity levels and introduce localized changes to scene content or road geometry~\cite{panacea2024,drivedreamer2,weatheredit2025,vace2025,qian2024weathergs,dai2025rainygs,sang2025weathermagician}. Although these approaches provide increasingly flexible visual control, their edits are generally decoupled from the corresponding vehicle response. For example, a model may render a wet or snow-covered road, or insert a speed hump or road depression, while retaining a prescribed ego trajectory that does not reflect the resulting changes in tire--road friction, braking distance, wheel slip, and the vehicle's vertical and pitching response. Conversely, vehicle simulators can compute such responses, but their simulated states are typically disconnected from photorealistic reconstructed scenes. The remaining gap is therefore a unified mechanism that maps the same weather
and road controls to both visual scene changes and the corresponding vehicle response (Fig.~\ref{fig:conceptual_gap}).

In this paper, we present muSync-GS, a physics-synchronized framework for driving-video synthesis under adverse-weather and road-elevation hazards. Unlike appearance-only editing, muSync-GS derives scene appearance, vehicle response, and ego-camera motion from shared weather and road controls. 

Our main contributions are summarized as follows:
\begin{itemize}
    \item We present muSync-GS, a physics-synchronized 3D Gaussian framework for controllable driving-video synthesis across rain and snow conditions and editable road-elevation hazards, including speed humps and road depressions. The framework generates the corresponding friction-dependent vehicle response and ego-camera motion.

    \item We construct a unified visual--physical pipeline that synthesizes road-surface appearance and dynamic rain or snow effects from precipitation inputs, while deriving a road-surface state that determines tire--road friction and
influences the resulting vehicle response. A shared road-elevation profile defines both the visible road geometry and vertical excitation. Using these friction and road inputs, a coupled longitudinal--vertical vehicle model predicts speed, slip ratio, dynamic normal loads, and pitch.
    \item We evaluate muSync-GS through frozen-parameter CarSim validation, real-vehicle brake-dive analysis on Audi Autonomous Driving Dataset (A2D2), and video-level synchronization and controllability experiments against video baselines on  reconstructed Waymo sequences.
\end{itemize}

\section{Related Work}

\paragraph{Controllable driving video generation and editing.}
Controllable driving-video generation has progressed from
structure-conditioned synthesis to temporally coherent multi-camera world modeling, supporting control through BEV layouts, maps, object boxes, camera parameters, actions, and trajectories~\cite{drivingdiffusion2024,magicdrive2024,wovogen2024,gaia1,panacea2024,drivedreamer2,gaia2_2025,unimlvg2025,safemvdrive2025,challenger2025,dtdima2025,ttcx2024}. Recent foundation models and video editors further support localized editing
of existing footage while preserving scene structure and temporal consistency~\cite{cosmostransfer2025,scenecrafter2025,vace2025,seeunsafe2025,valik2025}. In parallel, adverse-weather generation has progressed from global appearance transfer to controllable dynamic precipitation and cumulative effects such as snow accumulation and road-surface water buildup~\cite{lin2025weatherweaver}. However, these methods primarily treat weather as a visual condition. muSync-GS instead maps weather controls to a shared road-surface state that jointly determines scene appearance and vehicle response.

\paragraph{Editable 3D Gaussian driving scenes.}
3D Gaussian driving representations enable photorealistic reconstruction, generative 4D scene modeling, trajectory replay, sensor rendering, and semantic scene editing by modeling static environments and dynamic actors~\cite{yan2024streetgaussians,zhou2024drivinggaussian,zhou2024hugsim,splatad2024,horizonforge2026,streetunveiler2025,dreamdrive2025,drivedreamer4d2024,worldsplat2025}. Weather-aware extensions further support controllable dynamic precipitation, wet-road reflectance, and spatially accumulated snow~\cite{qian2024weathergs,dai2025rainygs,weatheredit2025,autoweather4d2026,sang2025weathermagician}. However, these weather effects are
modeled primarily for visual rendering: water films and snow layers alter scene appearance but are not linked to tire--road interaction or vehicle response. muSync-GS instead uses precipitation-induced road conditions as shared variables for both Gaussian appearance editing and friction-dependent vehicle simulation.

\paragraph{Physics-aware simulation and visual--physical coupling.}
Physics-integrated Gaussian methods combine 3D Gaussians with physical solvers, material properties, or learned force fields to model deformation and short-horizon dynamics~\cite{physgaussian2024,featuresplatting2024,dreamphysics2025,physics3d2024,physsplat2025,ngff2026}. Driving-oriented extensions further incorporate collision simulation or simulator-based vehicle control~\cite{huang2026real2sim,huang2026carlags}. RoVES applies a 4-DOF half-car model to road-profile edits, focusing on geometry-induced vertical displacement and pitch in reconstructed Gaussian scenes~\cite{physicsawaregs2026}. However, it does not model how weather-induced road-surface changes affect tire--road friction, longitudinal braking, or the resulting ego-vehicle trajectory. Physics-aware video-generation methods instead condition generative models on simulated motion, physical annotations, 4D occupancy, or externally updated scene states~\cite{physgen2024,vlipp2025,wisa2025,pathak2026physvid,geniedrive2025,pointasskeleton2026}. These methods mainly address material deformation, short-horizon object motion, or simulator-driven motion, while dedicated vehicle simulators remain visually disconnected from reconstructed real-world scenes~\cite{scenefactory2026,carla2017,carsim}. In contrast, muSync-GS jointly links weather and road-elevation controls to scene appearance, tire--road interaction, coupled longitudinal--vertical vehicle response, and ego-camera motion.

\section{Method}
\begin{figure*}[t]
\centering
\includegraphics[width=\textwidth]{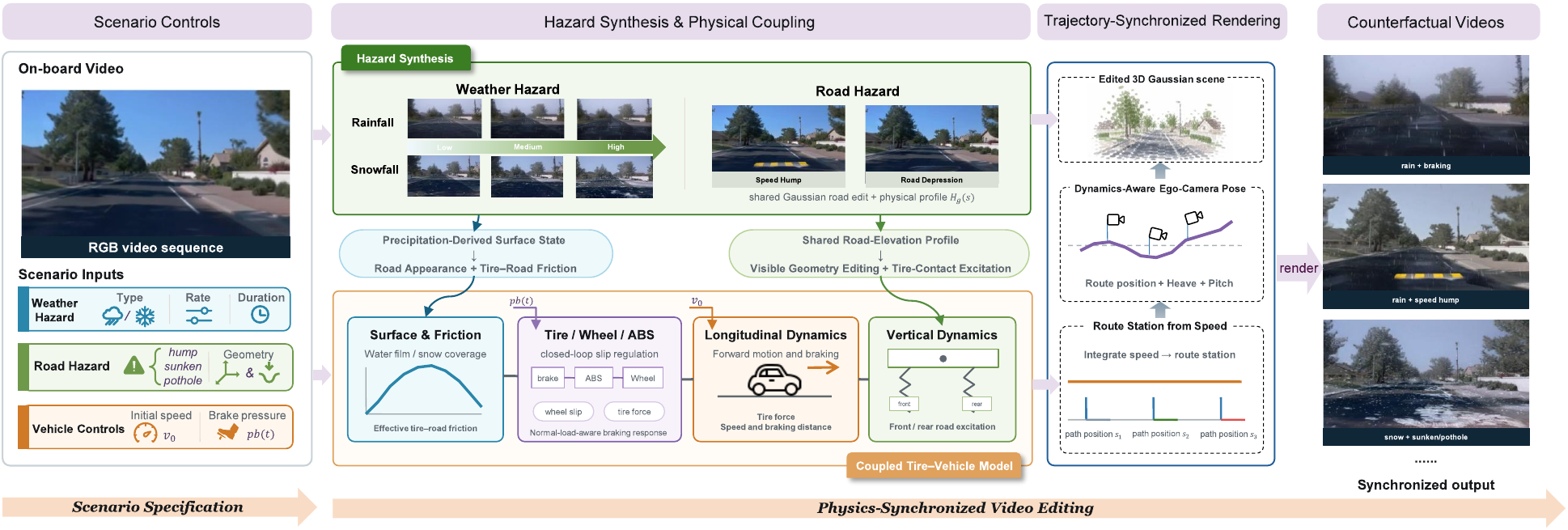}
\caption{Overview of muSync-GS. Weather, road-elevation, and vehicle controls
are converted into shared surface, friction, and geometry representations.
The muSync-GS vehicle model predicts speed, path position, wheel slip
ratios, symmetric wheel normal loads, and pitch. These states update the route
position and ego-camera pose for trajectory-synchronized rendering, producing
edited videos with frame-wise physical annotations.}
\label{fig:pipeline}
\end{figure*}
\subsection{Overview}

Given a reconstructed driving scene represented by 3D Gaussians
$\mathcal{G}$ and a reference ego trajectory $\mathcal{P}_0$, muSync-GS
generates a counterfactual video $\mathcal{V}$ and frame-synchronized physical
telemetry $\mathcal{Y}$. The scenario controls comprise weather type $w$,
precipitation rate $r$, snow-accumulation time $\tau_s$, initial speed $v_0$,
brake pressure $p_b(t)$, and a geometric road hazard represented by a
road-elevation profile with center position along the route $s_g$,
longitudinal extent $L_g$, and height $A_g$:
\begin{equation}
\begin{aligned}
(\mathcal{V},\mathcal{Y})
&=\mathcal{F}(\mathcal{G},\mathcal{P}_0,\mathbf{q},\mathbf{u},\mathbf{g}),\\
\mathbf{q}&=(w,r,\tau_s),\quad
\mathbf{u}=(v_0,p_b(t)),\\
\mathbf{g}&=(s_g,L_g,A_g).
\end{aligned}
\label{eq:overall_pipeline}
\end{equation}
Here, $r$ denotes the rainfall rate for rain and the water-equivalent
snowfall rate for snow. The variables $s_g$, $L_g$, and $A_g$ denote the
route-coordinate center, longitudinal length, and height of the road-elevation
profile, respectively. Positive and negative values of $A_g$ represent a speed
hump and a road depression, respectively, while $A_g=0$ corresponds to a flat
road. The vehicle motion and dynamically corrected ego-camera trajectory are
determined by the simulated response rather than prescribed independently.

\begin{table}[t]
\centering
{%
\small
\setlength{\tabcolsep}{3.5pt}
\begin{tabular}{@{}p{0.27\columnwidth}p{0.66\columnwidth}@{}}
\toprule
Notation & Meaning \\
\midrule
$\mathbf q,\mathbf u,\mathbf g$ & Weather, vehicle-control, and road-geometry controls. \\
$\mathbf z_{\mathrm{surf}}$ & Precipitation-derived surface state $(h_t,h_s,c_s)$. \\
$\mu_c,\mu_{\mathrm{eff}}$ & Reference surface coefficient and effective tire-curve peak. \\
$\boldsymbol\xi_{\mathrm{road}},\boldsymbol\xi_{\mathrm{atm}}$ & Road-appearance and atmospheric/particle rendering controls. \\
$H_g(s),s(t)$ & Signed road-elevation profile and front-axle route coordinate. \\
$\kappa_a,F_{x,a},F_{z,a}$ & Braking slip and axle-level tire force/load; wheel load is $F_{z,a}/2$ per side. \\
$\mathbf x_h$ & Half-car coordinates $(z_s,\theta,z_{u,f},z_{u,r})$. \\
$\mathcal P_0,\mathcal P(t)$ & Reference route pose and dynamically corrected camera pose. \\
\bottomrule
\end{tabular}
}
\caption{Principal notation used in the method. Subscripts $f$ and $r$ denote
the front and rear axles, respectively.}
\label{tab:method_parameters}
\end{table}

The framework is organized around three shared interfaces. A
precipitation-derived surface state connects weather to both road appearance
and tire--road friction. A shared road-elevation profile connects visible road
geometry to the height and slope experienced at the tire contacts. Finally,
the simulated path position connects the vehicle response to the rendered
ego-camera trajectory. These interfaces make appearance, geometry, vehicle
dynamics, and camera motion consequences of a single scenario specification.
Figure~\ref{fig:pipeline} summarizes the complete generation pipeline.

\subsection{Shared Visual--Physical Scene Controls}

\paragraph{Precipitation-derived surface state.}
We map the weather controls to
\begin{equation}
\mathbf{z}_{\mathrm{surf}}
=
\Phi_{\mathrm{surf}}(w,r,\tau_s)
=
(h_t,h_s,c_s),
\label{eq:surface_state}
\end{equation}
where $h_t$ denotes the total water-film depth, and $h_s$ and $c_s$ denote
snow depth and coverage.
Variables outside the selected weather family are set to zero. Rain uses a
steady-state drainage relation, while snow converts water-equivalent
precipitation into accumulated depth:
\begin{equation}
\begin{aligned}
h_t
&=
a\,\mathrm{MTD}^{b_1}L^{b_2}r^{b_3}S^{-b_4},\\
h_s
&=
r\tau_s\rho_w/\rho_s,\\
c_s
&=
\min(1,h_s/h_{\mathrm{cover}}).
\end{aligned}
\label{eq:surface_evolution}
\end{equation}
Here, $a,b_{1:4}$ are the Gallaway drainage coefficients; MTD, $L$, and
$S$ are the texture depth, drainage-path length, and crossfall; and
$\rho_w$, $\rho_s$, and $h_{\mathrm{cover}}$ are the water density, snow
density, and full-coverage depth, respectively. Their numerical values and
sources are reported in the supplementary material. The shared surface state
parameterizes the friction and rendering branches:
\begin{equation}
\begin{aligned}
\mu_c
&=
\phi_\mu(w,\mathbf{z}_{\mathrm{surf}}),\\
\boldsymbol{\xi}_{\mathrm{road}}
&=
\psi_{\mathrm{road}}(w,\mathbf{z}_{\mathrm{surf}}),\\
\boldsymbol{\xi}_{\mathrm{atm}}
&=
\psi_{\mathrm{atm}}(w,r).
\end{aligned}
\label{eq:condition_controls}
\end{equation}
The friction branch continuously maps the surface state to a reference friction coefficient:
\begin{equation}
\mu_c
=
\begin{cases}
\mu_{\mathrm{dry}},
& w=\mathrm{dry},\\
\mu_{\mathrm{dry}}
\left[
r_f+(1-r_f)e^{-h_t/h_c}
\right],
& w=\mathrm{rain},\\
\mu_{\mathrm{bw}}
-
(\mu_{\mathrm{bw}}-\mu_{\mathrm{snow}})c_s,
& w=\mathrm{snow}.
\end{cases}
\label{eq:surface_friction_mapping}
\end{equation}
Here, $r_f=\mu_{\mathrm{flooded}}/\mu_{\mathrm{dry}}$, $h_c$ is the
film-decay scale, $\mu_{\mathrm{bw}}$ is the bare-wet pavement anchor, and $\mu_{\mathrm{snow}}$ is the full snow-contact anchor. All anchors and scales are fixed before holdout evaluation; values and sensitivity analyses are provided in the supplement.


\paragraph{Unified road-elevation profile.}
The geometry control defines a raised-cosine height field. Let
$\delta_g=s-s_g$ denote the offset from the profile center:
\begin{equation}
H_g(s)
=
\begin{cases}
\displaystyle
\frac{A_g}{2}
\left[
1+\cos\left(
\frac{2\pi\delta_g}{L_g}
\right)
\right],
& |\delta_g|\leq L_g/2,\\
0,
& \text{otherwise}.
\end{cases}
\label{eq:shared_road_profile}
\end{equation}
The same $H_g$ defines the visible 3D road-geometry edit and supplies the
height and slope at the front and rear tire contacts. The shared profile therefore synchronizes the visible road edit with the excitation applied at the tire contacts.

\paragraph{Conditioned scene synthesis.}
Atmospheric appearance, road accumulation, and falling particles are rendered
as separate components from the corrected camera pose.
WeatherEdit~\cite{weatheredit2025} provides the atmospheric transformation
and particle render. We isolate the dynamic particle component using paired
renders:
\begin{equation}
\Delta I_{\mathrm{part}}(t)
=
I_{\mathrm{with\mbox{-}part}}(t)
-
I_{\mathrm{no\mbox{-}part}}(t),
\label{eq:particle_residual}
\end{equation}
where the two image terms are pose-matched renders with and without falling
precipitation. Their residual $\Delta I_{\mathrm{part}}$ is composited after
pose synchronization. Rain appearance uses road semantics, depth, and surface normals, following the reflection, refraction, and wet-surface compositing formulation of RainyGS~\cite{dai2025rainygs}. Snow accumulation adapts the
normal-guided principle of
Weather-Magician~\cite{sang2025weathermagician}: road-surface anchors seed a
nested appended Gaussian snow layer. Coverage $c_s$ selects the active subset
through a deterministic coverage field, while the selected Gaussians are
displaced along the local road normal according to $h_s$. Detailed anchor
construction, interpolation rules, and rendering parameters are provided in the
supplementary material.
\subsection{Coupled Vehicle Response}

The surface coefficient sets the peak of a weather-family-specific tire curve.
For axle $a\in\{f,r\}$, braking slip and longitudinal tire force are
\begin{equation}
\begin{aligned}
\mu_{\mathrm{eff}}&=\eta_{\mathrm{tire}}(w,\mu_c)\mu_c,\\
\kappa_a&=\operatorname{clip}\!\left(
\frac{v-R\omega_a}{\max(v,0.5)},0,1\right),\\
F_{x,a}&=\mu_{\mathrm B}
(\beta_{\kappa,w}\kappa_a;\mu_{\mathrm{eff}})F_{z,a},
\end{aligned}
\label{eq:tire_force}
\end{equation}
Here $\eta_{\mathrm{tire}}$ is the frozen weather-dependent scaling factor,
$\mu_{\mathrm B}$ is the dry, wet, or snow Burckhardt curve normalized to
$\mu_{\mathrm{eff}}$, and $\beta_{\kappa,w}$ is the corresponding frozen
slip-stiffness scale. Brake pressure and ABS dynamics determine $\omega_a$;
their controller equations are provided in the supplementary material.

Here $s(t)$ denotes the front-axle route coordinate, with $\dot{s}=v$,
$z_{r,f}=H_g(s)$, and $z_{r,r}=H_g(s-L_w)$. The shared road-elevation profile
contributes at both axle contacts to the longitudinal dynamics:
\begin{equation}
\begin{aligned}
m\dot v
&=-(F_{x,f}+F_{x,r})-F_{z,f}H'_g(s)\\
&\quad-F_{z,r}H'_g(s-L_w)-ma_{\mathrm{coast}},
\qquad \dot{s}=v.
\end{aligned}
\label{eq:longitudinal_dynamics}
\end{equation}
Here $L_w$ is the wheelbase and $a_{\mathrm{coast}}$ is the coast-drag
deceleration. Road and braking excitation enter a common 4-DOF half-car model,
\begin{equation}
\mathbf M_h\ddot{\mathbf x}_h+
\mathbf C_h\dot{\mathbf x}_h+
\mathbf K_h\mathbf x_h
=\mathbf B_r\mathbf z_r+\mathbf B_bM_b,
\label{eq:half_car}
\end{equation}
where $\mathbf x_h=(z_s,\theta,z_{u,f},z_{u,r})$ and
$\mathbf z_r=(z_{r,f},z_{r,r})^\top$. The model returns dynamic front/rear
normal loads, body heave, and pitch. The loads feed back into the longitudinal
tire forces, while heave and pitch update the rendered camera pose. Mechanical
parameters are fixed from the matched vehicle; the remaining parameters are
calibrated on the 19-case development set and frozen before evaluation.
Controller, contact, low-speed, coupled-update, and calibration details are
provided in the supplementary material.

\subsection{Trajectory-Synchronized Rendering and Outputs}

Integrating the simulated speed gives the longitudinal path position $s(t)$,
i.e., the distance traveled along the reference route. We use $s(t)$ to continuously sample the reference route. Body heave and pitch then
correct the reference pose:
\begin{equation}
\mathcal P(t)=\mathcal P_0(s(t))\oplus(z_s(t),\theta(t)).
\label{eq:camera_pose}
\end{equation}
The shared road-elevation profile first determines the visible
road-geometry edit, after which the
atmospheric and accumulated-road layers are rendered from $\mathcal P(t)$.
The pose-matched particle residual from Eq.~\ref{eq:particle_residual} is added
last. After the vehicle stops, the static scene remains fixed while precipitation
particles continue to evolve.

Each output frame is accompanied by surface state, surface coefficient $\mu_c$,
tire-curve peak $\mu_{\mathrm{eff}}$, speed, path position, acceleration, wheel slip, dynamic normal
loads, heave, pitch, and brake pressure.

\section{Experiments}
\label{sec:experiments}
We evaluate muSync-GS along four dimensions: frozen-parameter vehicle-response accuracy against CarSim cases, real-vehicle brake-dive consistency on A2D2, video-motion synchronization and visual controllability on reconstructed Waymo sequences, and the contribution of individual visual--physical couplings through ablations.
\subsection{Experimental Setup}
\label{sec:experimental_setup}

We evaluate two scenario dimensions: weather-dependent road conditions and
geometric road hazards. We evaluate a dry-weather baseline and rain and snow conditions at multiple precipitation levels. The road-elevation settings include a flat-road baseline, a speed hump, and a road depression. Rain and snow modify tire--road friction and braking response, whereas the speed hump and road depression excite the axle and suspension dynamics.

We characterize the vehicle response using longitudinal speed, wheel slip
ratio, dynamic normal load, and body pitch. At the video level,
we evaluate ego forward progress and camera pitch recovered from the rendered
sequence. 

Experiments use Waymo Open Dataset sequences~\cite{sun2020waymo}. The 19-case
development set contains nine flat-road braking, five speed-hump, and five
road-depression cases. The 12-case CarSim evaluation covers held-out intermediate rain and snow severity levels, brake commands, and road-profile heights and longitudinal extents. With all calibrated parameters frozen, we compare speed, body pitch, slip ratio, and per-wheel normal load on the original time axis, without temporal alignment or parameter refitting.

For the external real-vehicle evaluation, we detect 140 brake-pressure episodes
in three complete A2D2 Audi Q7 drives~\cite{geyer2020a2d2}; 138 support the full
analysis window and 14 satisfy the straight-line and signal-quality criteria
($8/2/4$ events; initial speeds $13.1$--$36.4\,\mathrm{km/h}$). We perform
complete-drive leave-one-drive-out evaluation over the first $0.5\,\mathrm{s}$
after brake-pressure onset. Vehicle-specific parameters and the pressure gain
are estimated from the two training drives and frozen for the held-out drive.

VGGT~\cite{wang2025vggt} recovers normalized forward progress and camera pitch from each generated video for comparison with CarSim responses. We also evaluate lateral camera-path preservation after aligning trajectories by normalized forward progress. Results for stochastic video generators are averaged over three seeds.

\subsection{Vehicle and Video Response}
\paragraph{Frozen-Parameter Vehicle-Response Validation against CarSim.}

\begin{table}[!t]
\centering
{%
\small
\setlength{\tabcolsep}{1.4pt}
\begin{tabular}{@{}lrrrr@{}}
\hline
Set (cases) & Speed & Pitch & Slip & $F_z$ \\
\hline
Weather (5)  & .044/.099 & .066/.102 & .0207/.0367 & 22.3/37.0 \\
Brake (3)    & .021/.024 & .074/.095 & .0057/.0059 & 28.3/36.0 \\
Road (4) & .011/.011 & .039/.053 & .00007/.00011 & 30.8/37.7 \\
\hline
\end{tabular}

\vspace{3pt}
\begin{tabular}{@{}llr@{}}
\hline
Set & Event metric & Error mean/max \\
\hline
Weather & Stop distance (m) & 0.141/0.284 \\
Brake & Stop distance (m) & 0.0183/0.0210 \\
Road profile & Center speed (km/h) & 0.0391/0.0406 \\
\hline
\end{tabular}
}
\caption{Frozen-parameter agreement on 12 held-out CarSim cases. Entries are
case-level mean / maximum; state RMSE units are m/s, degrees, unitless slip
ratio, and N.}
\label{tab:carsim_validation}
\end{table}

\begin{table*}[!t]
\centering
{%
\small
\setlength{\tabcolsep}{4.0pt}
\begin{tabular}{lccccc}
\hline
& Weather response
& \multicolumn{2}{c}{Geometry response}
& Motion preservation & Path preservation \\
\cline{2-2}\cline{3-4}\cline{5-5}\cline{6-6}
Method & \shortstack{Norm. progress\\RMSE $\downarrow$}
& \shortstack{Hump\\NRMSE $\downarrow$}
& \shortstack{Depression\\NRMSE $\downarrow$}
& \shortstack{Abs. pitch corr.\\$\uparrow$}
& \shortstack{Norm. lateral\\RMS $\downarrow$} \\
\hline
Cosmos & $0.0862\!\pm\!0.0022$
       & $1.003\!\pm\!0.009$ & $1.006\!\pm\!0.007$
       & $\mathbf{0.996}\!\pm\!0.001$ & $0.0036\!\pm\!0.0011$ \\
LTX-Video & $0.1269\!\pm\!0.0579$
       & $0.992\!\pm\!0.027$ & $1.003\!\pm\!0.019$
       & $0.988\!\pm\!0.005$ & $0.0126\!\pm\!0.0054$ \\
VACE & $0.0867\!\pm\!0.0050$
       & $0.999\!\pm\!0.003$ & $1.000\!\pm\!0.002$
       & $0.995\!\pm\!0.002$ & $0.0035\!\pm\!0.0006$ \\
muSync-GS & $\mathbf{0.00424\!\pm\!0.00010}$
       & $\mathbf{0.316}^{\mathrm{det.}}$ & $\mathbf{0.431}^{\mathrm{det.}}$
       & -- & $\mathbf{0.00266\!\pm\!0.00007}$ \\
\hline
\end{tabular}
}
\caption{VGGT video-motion synchronization against CarSim. All metrics
are dimensionless; stochastic entries report mean $\pm$ standard deviation
over three seeds, and \emph{det.} denotes deterministic results.}
\label{tab:vggt_motion}
\end{table*}

\begin{figure}[!t]
\centering
\includegraphics[width=0.85\columnwidth,height=0.72\columnwidth]
{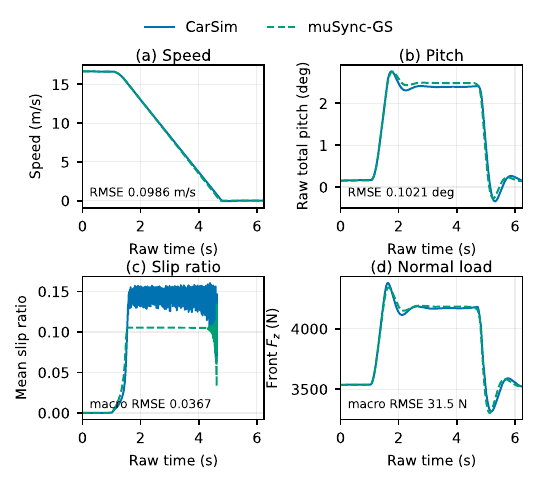}
\caption{muSync-GS and CarSim trajectories for a held-out
$5\,\mathrm{mm/h}$ rain braking case: speed, pitch, mean wheel slip ratio,
and front-wheel normal load.}
\label{fig:carsim_ood_representatives}
\end{figure}

Across all cases, macro-averaged RMSEs are $0.0273\,\mathrm{m/s}$ for speed,
$0.0590^\circ$ for pitch, $0.0101$ for slip, and $26.61\,\mathrm{N}$ for
normal load.
\paragraph{Real-vehicle brake-dive evaluation.}
We further evaluate whether vehicle-specific calibration enables muSync-GS to
capture the initial brake-dive response in real-vehicle data. For each event in
a held-out drive, the model receives only the brake-pressure trace and the
vehicle speed at event onset. All calibration parameters are estimated from
the remaining drives. Across 14 braking events from three A2D2 drives, the
calibrated model achieves drive-macro-averaged pitch and pitch-rate RMSEs of
$0.062^\circ$ and $0.381^\circ/\mathrm{s}$, respectively, over the initial
$0.5$~s, with corresponding pitch and pitch-rate correlations of $0.901$ and
$0.560$.

\begin{figure}[!t]
\centering
\includegraphics[width=0.85\columnwidth]
{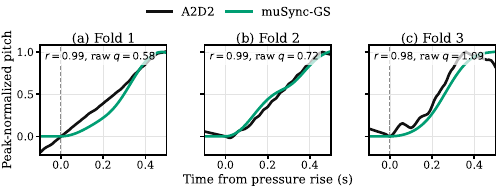}
\caption{Peak-magnitude-normalized measured and predicted brake-dive responses from each held-out A2D2 drive. The corresponding unnormalized
peak ratio $q_\theta$ is reported in each panel.}
\label{fig:a2d2_lodo_examples}
\end{figure}

Figure~\ref{fig:a2d2_lodo_examples} illustrates the temporal agreement between
measured and predicted pitch for one event from each held-out drive. For
visualization, each trajectory is normalized by its own peak magnitude, while
the corresponding unnormalized peak ratio $q_\theta$ is reported in each panel.
This evaluation provides an additional real-vehicle check of the initial
brake-dive response after vehicle-specific calibration.

\begin{figure}[t]
\centering
\includegraphics[width=\columnwidth]
{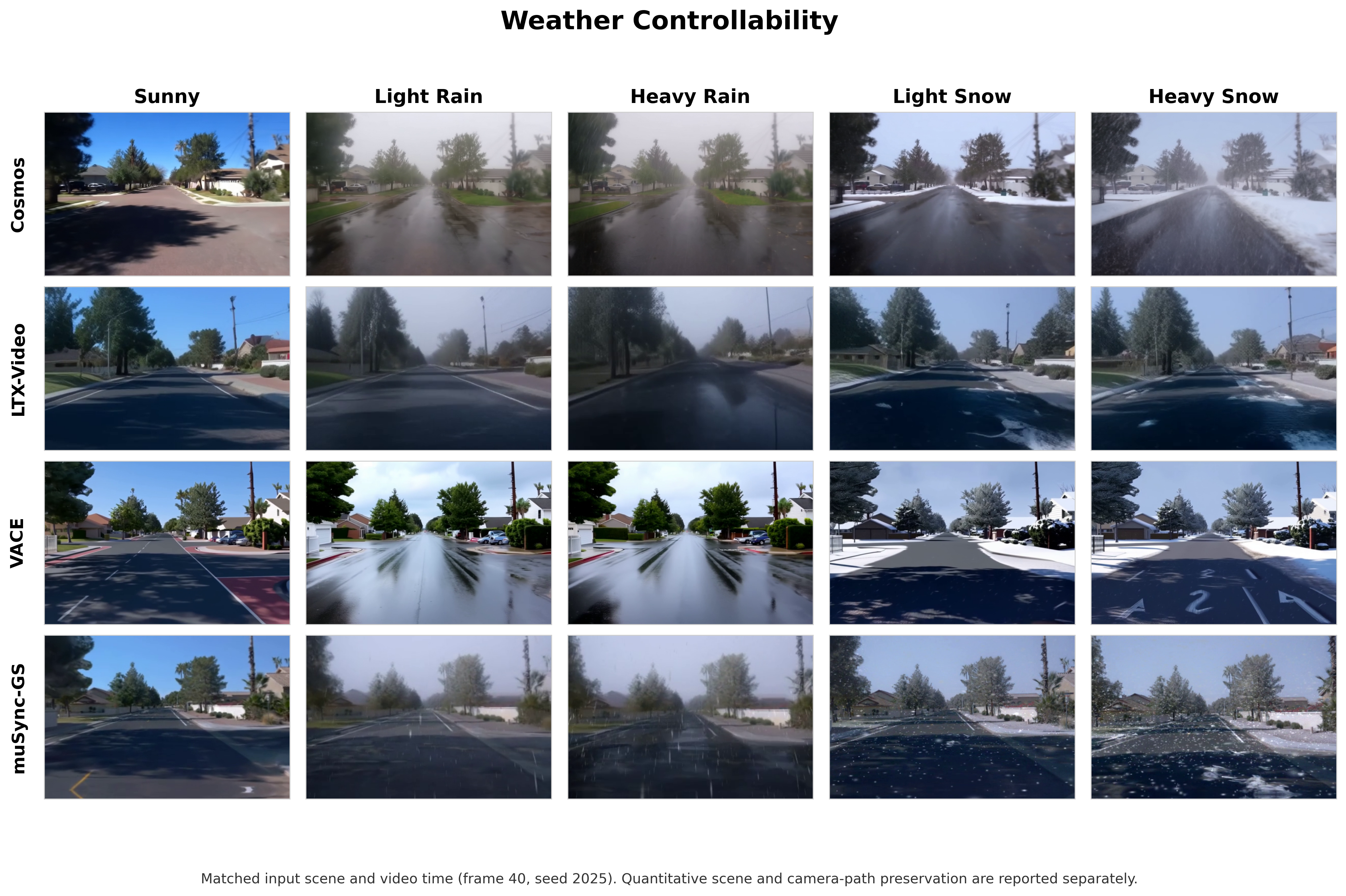}
\caption{Five-level weather controllability at frame 40 and seed 2025: sunny,
light rain ($10\,\mathrm{mm/h}$), heavy rain ($50\,\mathrm{mm/h}$), light snow
($1\,\mathrm{mm/h}$ water equivalent), and heavy snow ($3\,\mathrm{mm/h}$
water equivalent). Every row uses a matched source scene and video time.}
\label{fig:weather_controllability}
\end{figure}

\begin{figure*}[!t]
\centering
\includegraphics[width=\textwidth]
{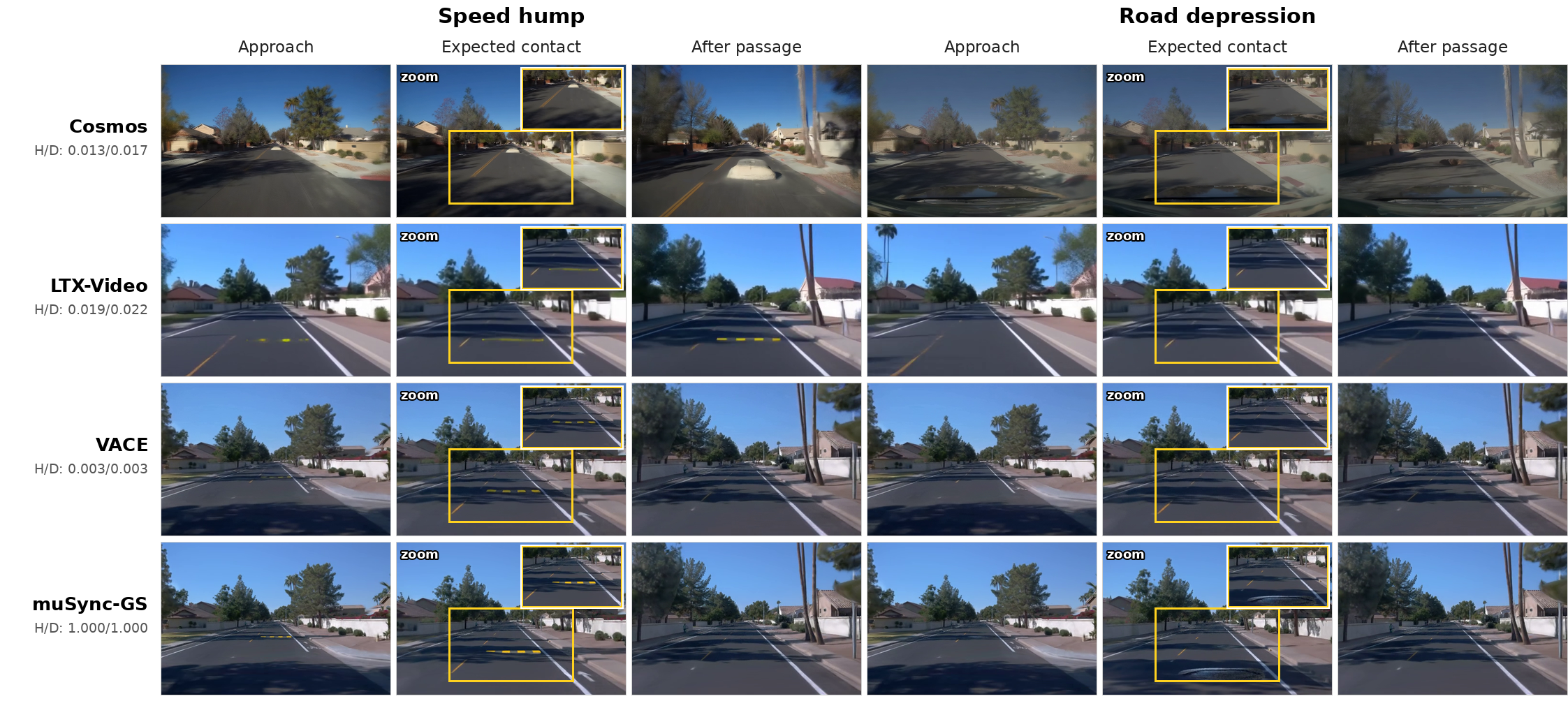}
\caption{Road editing under the same dry-weather input (seed 2025) before contact, at front-axle contact, and after passage. Insets enlarge the contact region; H/D
labels report the hump/depression response from Table~\ref{tab:vggt_motion}.}
\label{fig:road_geometry_comparison}
\end{figure*}

\paragraph{Video-Response Synchronization against Video Baselines.}
We compare muSync-GS with Cosmos, LTX-Video, and VACE over three random seeds.
VGGT recovers normalized ego forward progress and camera pitch from each
generated video. We evaluate whether the edited videos reproduce the
weather-dependent braking response and the camera-pitch response induced by a
speed hump or road depression, using the corresponding CarSim trajectories as
references. We separately measure how well each method preserves motion already
present in the input video. All comparisons use fixed common-time windows,
without method-specific temporal alignment or fitting.

muSync-GS achieves hump and depression NRMSEs of $0.316$ and $0.431$ respectively. The baselines remain
near the no-response geometry score and show substantially larger
weather-response error, although input--output correlations of
$0.988$--$0.996$ show that they can preserve motion already present in the
input. Thus, the evaluation separates motion preservation from inference of missing
weather- or geometry-dependent dynamics. Weather-response RMSE uses normalized
progress because monocular translation has unknown scale. The
final column shows that muSync-GS also has the lowest progress-aligned lateral
camera-path deviation.

\subsection{Visual Quality, Controllability, and Consistency}
\label{sec:visual_results}
\paragraph{Weather appearance and severity control.}
Figure~\ref{fig:weather_controllability} compares weather appearance at the same scene and frame. For automatic appearance
evaluation, we report the overall score of the no-reference DOVER
evaluator~\cite{wu2023dover}. We additionally measure semantic severity
alignment with CLIP~\cite{radford2021clip}: 16 uniformly sampled frames are
scored against prompt ensembles for dry, heavy-rain, and heavy-snow dashcam
scenes, and Spearman's $\rho$ is computed between the continuous precipitation
input and the corresponding weather-minus-dry similarity. DOVER measures
generic perceptual video quality, whereas CLIP $\rho$ measures control
monotonicity rather than photorealism. VACE obtains the highest DOVER score; however,
adverse-weather effects themselves, including reduced visibility,
precipitation particles, reflections, and snow occlusion, can lower
no-reference quality scores. Consequently, methods that make milder edits and
retain more of the clear source video have an inherent advantage under this
metric, even when the requested weather effect is less pronounced.

\paragraph{Road-geometry editing.}
Paired feature and flat-control videos measure spatial localization of the road edit and
optical-flow-warped temporal mask IoU (T-IoU). Missing detections are
penalized; full metric definitions are provided in the supplement.

\begin{table}[!t]
\centering
{%
\small
\setlength{\tabcolsep}{3.0pt}
\begin{tabular}{lccc}
\hline
Method & \shortstack{DOVER\\Overall $\uparrow$}
& \shortstack{Rain\\$\rho\uparrow$}
& \shortstack{Snow\\$\rho\uparrow$} \\
\hline
Cosmos & $0.176\!\pm\!0.017$ & $0.967\!\pm\!0.058$ & $0.867\!\pm\!0.058$ \\
LTX-Video & $0.203\!\pm\!0.037$ & $0.933\!\pm\!0.058$ & $0.867\!\pm\!0.231$ \\
VACE & $\mathbf{0.368\!\pm\!0.027}$ & $0.933\!\pm\!0.058$ & $0.667\!\pm\!0.351$ \\
muSync-GS & $0.180\!\pm\!0.002$ & $\mathbf{1.000\!\pm\!0.000}$ & $\mathbf{1.000\!\pm\!0.000}$ \\
\hline
\end{tabular}

\vspace{3pt}
\begin{tabular}{lcc}
\hline
Method & \shortstack{Localized edit\\rate $\uparrow$} & T-IoU $\uparrow$ \\
\hline
Cosmos & $0.338\!\pm\!0.158$ & $0.253\!\pm\!0.162$ \\
LTX-Video & $0.303\!\pm\!0.086$ & $0.114\!\pm\!0.049$ \\
VACE & $0.570\!\pm\!0.050$ & $0.335\!\pm\!0.006$ \\
muSync-GS & $\mathbf{0.711}^{\mathrm{det.}}$ & $\mathbf{0.603}^{\mathrm{det.}}$ \\
\hline
\end{tabular}
}
\caption{Dimensionless weather/road visual metrics. Stochastic entries report
mean $\pm$ standard deviation over three seeds; \emph{det.} denotes
deterministic results.}
\label{tab:geometry_visual_consistency}
\end{table}
muSync-GS preserves the expected severity ordering for both weather families and achieves the highest localized-edit and T-IoU scores. These results show that the proposed controls produce monotonic weather changes and localized, temporally consistent road edits. Figure~\ref{fig:road_geometry_comparison} further shows that the baseline edits are less spatially localized and temporally consistent. Together with the geometry-response evaluation, this comparison
distinguishes visible road editing from synchronized vehicle response.


\subsection{Ablation Study}
\label{sec:ablations}
\paragraph{Ablation variants.}
We compare Appearance-only, Vertical-only, Fixed-$\mu$, Full muSync-GS, and a
CarSim-pose oracle. Vertical-only retains the road-elevation profile and
half-car dynamics while disabling weather-dependent friction and braking
response. Fixed-$\mu$ replaces continuous surface
state with a weather-family lookup. The CarSim-pose oracle renders CarSim poses and serves as an oracle reference. All wheel-dynamics
variants use the same frozen vehicle, tire, brake, and ABS parameters matched to CarSim;
the ablations change only the coupling identified by their labels.

\begin{figure}[!t]
\centering
\includegraphics[width=0.85\columnwidth]
{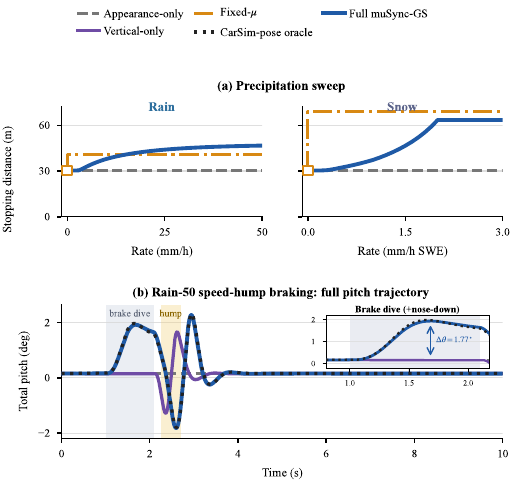}
\caption{Coupled-physics ablation: (a) stopping distance across rain and snow
severity; (b) pitch under $50\,\mathrm{mm/h}$ rain braking and speed-hump traversal. Shaded
regions mark braking and front-axle traversal.}
\label{fig:weather_pitch_ablation}
\end{figure}

\paragraph{Weather--friction coupling.}
We evaluate densely sampled rain intensities from
$0$ to $50\,\mathrm{mm/h}$ and water-equivalent snowfall rates from
$0$ to $3\,\mathrm{mm/h}$ at an initial speed of
$60\,\mathrm{km/h}$.
Appearance-only retains dry dynamics, and Fixed-$\mu$ changes discontinuously
from dry to one constant response per weather family. The full model instead produces
continuous severity-dependent stopping distance and explicit saturation after
full snow cover (Fig.~\ref{fig:weather_pitch_ablation}(a)).

\paragraph{Full pitch trajectory under coupled braking and geometry.}
Vertical-only responds only to the road profile, whereas full
muSync-GS exhibits brake dive before reaching the hump, reaches the hump later under deceleration, and couples front- and
rear-axle excitation with the braking transient
(Fig.~\ref{fig:weather_pitch_ablation}(b)). The CarSim-pose oracle follows the
same response with $0.031^\circ$ full-trajectory RMSE.


\section{Conclusion}
muSync-GS couples weather and road edits to tire--vehicle response and ego-camera
motion through shared surface, geometry, and trajectory states. Frozen-parameter
experiments show synchronized visual and physical responses across held-out
weather, braking, and road controls. An external vehicle-adapted A2D2 evaluation
additionally supports consistency of the initial brake-dive trend. Together, these results demonstrate the feasibility of generating driving
counterfactuals in which visual hazards and ego motion remain physically
synchronized. Future work will incorporate measured
tire--road data, extend the framework to broader vehicle, tire, and road
conditions and lateral dynamics, and evaluate the generated counterfactuals
for autonomous-driving policy training and testing.

\begingroup
\fontsize{10.95pt}{12.6pt}\selectfont
\setlength{\bibsep}{1pt}
\bibliographystyle{unsrtnat}
\bibliography{musyncgs,AppendixReferences}

\begin{thebibliography}{68}
\providecommand{\natexlab}[1]{#1}
\providecommand{\url}[1]{\texttt{#1}}
\expandafter\ifx\csname urlstyle\endcsname\relax
  \providecommand{\doi}[1]{doi: #1}\else
  \providecommand{\doi}{doi: \begingroup \urlstyle{rm}\Url}\fi

\bibitem[Hu et~al.(2023)Hu, Russell, Yeo, Murez, Fedoseev, Kendall, Shotton,
  and Corrado]{gaia1}
Anthony Hu, Lloyd Russell, Hudson Yeo, Zak Murez, George Fedoseev, Alex
  Kendall, Jamie Shotton, and Gianluca Corrado.
\newblock Gaia-1: A generative world model for autonomous driving, 2023.

\bibitem[Wen et~al.(2024)Wen, Zhao, Liu, Jia, Wang, Luo, Zhang, Wang, Sun, and
  Zhang]{panacea2024}
Yuqing Wen, Yucheng Zhao, Yingfei Liu, Fan Jia, Yanhui Wang, Chong Luo, Chi
  Zhang, Tiancai Wang, Xiaoyan Sun, and Xiangyu Zhang.
\newblock Panacea: Panoramic and controllable video generation for autonomous
  driving.
\newblock In \emph{Proceedings of the IEEE/CVF Conference on Computer Vision
  and Pattern Recognition}, pages 6902--6912, 2024.
\newblock \doi{10.1109/CVPR52733.2024.00659}.

\bibitem[Zhao et~al.(2025{\natexlab{a}})Zhao, Wang, Zhu, Chen, Huang, Bao, and
  Wang]{drivedreamer2}
Guosheng Zhao, Xiaofeng Wang, Zheng Zhu, Xinze Chen, Guan Huang, Xiaoyi Bao,
  and Xingang Wang.
\newblock Drivedreamer-2: Llm-enhanced world models for diverse driving video
  generation.
\newblock In \emph{Proceedings of the AAAI Conference on Artificial
  Intelligence}, volume~39, pages 10412--10420, 2025{\natexlab{a}}.
\newblock \doi{10.1609/aaai.v39i10.33130}.

\bibitem[Li et~al.(2025)Li, Bian, Lei, Zuo, Yang, Zhu, Li, Chen, and
  Ozbay]{dtdima2025}
Tao Li, Zilin Bian, Haozhe Lei, Fan Zuo, Ya-Ting Yang, Quanyan Zhu, Zhenning
  Li, Zhibin Chen, and Kaan Ozbay.
\newblock Digital twin-based driver risk-aware predictive mobility analytics
  for real-time situational awareness through cooperative sensing.
\newblock \emph{IEEE Transactions on Intelligent Transportation Systems},
  26\penalty0 (11):\penalty0 20071--20090, 2025.
\newblock \doi{10.1109/TITS.2025.3604569}.

\bibitem[Zhang et~al.(2025)Zhang, Wang, Zhang, Bian, Feng, and
  Ozbay]{seeunsafe2025}
Ruixuan Zhang, Beichen Wang, Juexiao Zhang, Zilin Bian, Chen Feng, and Kaan
  Ozbay.
\newblock When language and vision meet road safety: Leveraging multimodal
  large language models for video-based traffic accident analysis.
\newblock \emph{Accident Analysis \& Prevention}, 219:\penalty0 108077, 2025.
\newblock \doi{10.1016/j.aap.2025.108077}.

\bibitem[Li et~al.(2024{\natexlab{a}})Li, Bian, Lei, Zuo, Yang, Zhu, Li, and
  Ozbay]{ttcx2024}
Tao Li, Zilin Bian, Haozhe Lei, Fan Zuo, Ya-Ting Yang, Quanyan Zhu, Zhenning
  Li, and Kaan Ozbay.
\newblock Multi-level traffic-responsive tilt camera surveillance through
  predictive correlated online learning.
\newblock \emph{Transportation Research Part C: Emerging Technologies},
  167:\penalty0 104804, 2024{\natexlab{a}}.
\newblock \doi{10.1016/j.trc.2024.104804}.

\bibitem[Liu et~al.(2025)Liu, Meng, Gao, Mao, Cai, Yan, Chen, Bian, Wang, and
  Shi]{valik2025}
Junming Liu, Siyuan Meng, Yanting Gao, Song Mao, Pinlong Cai, Guohang Yan,
  Yirong Chen, Zilin Bian, Ding Wang, and Botian Shi.
\newblock Aligning vision to language: Annotation-free multimodal knowledge
  graph construction for enhanced {LLMs} reasoning.
\newblock In \emph{Proceedings of the IEEE/CVF International Conference on
  Computer Vision}, pages 981--992, 2025.
\newblock \doi{10.1109/ICCV51701.2025.00099}.

\bibitem[Qian et~al.(2026)Qian, Li, Guo, and Markkula]{weatheredit2025}
Chenghao Qian, Wenjing Li, Yuhu Guo, and Gustav Markkula.
\newblock Weatheredit: Controllable weather editing with {4D} gaussian field.
\newblock In \emph{Proceedings of the AAAI Conference on Artificial
  Intelligence}, 2026.
\newblock \doi{10.1609/aaai.v40i10.37802}.
\newblock arXiv:2505.20471.

\bibitem[Jiang et~al.(2025)Jiang, Han, Mao, Zhang, Pan, and Liu]{vace2025}
Zeyinzi Jiang, Zhen Han, Chaojie Mao, Jingfeng Zhang, Yulin Pan, and Yu~Liu.
\newblock Vace: All-in-one video creation and editing.
\newblock In \emph{Proceedings of the IEEE/CVF International Conference on
  Computer Vision}, 2025.
\newblock \doi{10.1109/ICCV51701.2025.01597}.

\bibitem[Qian et~al.(2025)Qian, Guo, Li, and Markkula]{qian2024weathergs}
Chenghao Qian, Yuhu Guo, Wenjing Li, and Gustav Markkula.
\newblock {WeatherGS}: {3D} scene reconstruction in adverse weather conditions
  via gaussian splatting.
\newblock In \emph{IEEE International Conference on Robotics and Automation},
  2025.
\newblock \doi{10.1109/ICRA55743.2025.11128699}.
\newblock arXiv:2412.18862.

\bibitem[Dai et~al.(2025)Dai, Ni, Shen, Chen, Chen, and Chu]{dai2025rainygs}
Qiyu Dai, Xingyu Ni, Qianfan Shen, Wenzheng Chen, Baoquan Chen, and Mengyu Chu.
\newblock {RainyGS}: Efficient rain synthesis with physically-based gaussian
  splatting.
\newblock In \emph{Proceedings of the IEEE/CVF Conference on Computer Vision
  and Pattern Recognition}, 2025.
\newblock \doi{10.1109/CVPR52734.2025.01506}.

\bibitem[Sang et~al.(2025)Sang, Qian, Zhang, Wang, and
  Yang]{sang2025weathermagician}
Chen Sang, Yeqiang Qian, Jiale Zhang, Chunxiang Wang, and Ming Yang.
\newblock Weather-magician: Reconstruction and rendering framework for {4D}
  weather synthesis in real time, 2025.
\newblock URL \url{https://arxiv.org/abs/2505.19919}.

\bibitem[Li et~al.(2024{\natexlab{b}})Li, Zhang, and Ye]{drivingdiffusion2024}
Xiaofan Li, Yifu Zhang, and Xiaoqing Ye.
\newblock Drivingdiffusion: Layout-guided multi-view driving scenarios video
  generation with latent diffusion model.
\newblock In \emph{European Conference on Computer Vision}, 2024{\natexlab{b}}.
\newblock \doi{10.1007/978-3-031-73229-4_27}.

\bibitem[Gao et~al.(2024)Gao, Chen, Xie, Hong, Li, Yeung, and
  Xu]{magicdrive2024}
Ruiyuan Gao, Kai Chen, Enze Xie, Lanqing Hong, Zhenguo Li, Dit-Yan Yeung, and
  Qiang Xu.
\newblock Magicdrive: Street view generation with diverse 3d geometry control,
  2024.
\newblock URL \url{https://arxiv.org/abs/2310.02601}.

\bibitem[Lu et~al.(2024)Lu, Huang, Yang, Zhang, and Zhang]{wovogen2024}
Jiachen Lu, Ze~Huang, Zeyu Yang, Jiahui Zhang, and Li~Zhang.
\newblock Wovogen: World volume-aware diffusion for controllable multi-camera
  driving scene generation.
\newblock In \emph{European Conference on Computer Vision}, 2024.
\newblock \doi{10.1007/978-3-031-72989-8_19}.

\bibitem[Russell et~al.(2025)Russell, Hu, Bertoni, Fedoseev, Shotton, Arani,
  and Corrado]{gaia2_2025}
Lloyd Russell, Anthony Hu, Lorenzo Bertoni, George Fedoseev, Jamie Shotton,
  Elahe Arani, and Gianluca Corrado.
\newblock Gaia-2: A controllable multi-view generative world model for
  autonomous driving, 2025.
\newblock URL \url{https://arxiv.org/abs/2503.20523}.

\bibitem[Chen et~al.(2025)Chen, Wu, Liu, Guo, Ni, Xia, and Xia]{unimlvg2025}
Rui Chen, Zehuan Wu, Yichen Liu, Yuxin Guo, Jingcheng Ni, Haifeng Xia, and Siyu
  Xia.
\newblock Unimlvg: Unified framework for multi-view long video generation with
  comprehensive control capabilities for autonomous driving.
\newblock In \emph{Proceedings of the IEEE/CVF International Conference on
  Computer Vision}, 2025.
\newblock \doi{10.1109/ICCV51701.2025.02361}.

\bibitem[Zhou et~al.(2025)Zhou, Lyu, Tian, Zhuo, and Li]{safemvdrive2025}
Jiawei Zhou, Linye Lyu, Zhuotao Tian, Cheng Zhuo, and Yu~Li.
\newblock {SafeMVDrive}: Multi-view safety-critical driving video synthesis in
  the real world domain, 2025.
\newblock URL \url{https://arxiv.org/abs/2505.17727}.

\bibitem[Xu et~al.(2025{\natexlab{a}})Xu, Li, Gao, Gao, Chen, Liu, Yan, Zhao,
  Feng, and Zhao]{challenger2025}
Zhiyuan Xu, Bohan Li, Huan-ang Gao, Mingju Gao, Yong Chen, Ming Liu, Chenxu
  Yan, Hang Zhao, Shuo Feng, and Hao Zhao.
\newblock Challenger: Affordable adversarial driving video generation,
  2025{\natexlab{a}}.
\newblock URL \url{https://arxiv.org/abs/2505.15880}.

\bibitem[{NVIDIA}(2025)]{cosmostransfer2025}
{NVIDIA}.
\newblock Cosmos-transfer1: Conditional world generation with adaptive
  multimodal control, 2025.
\newblock URL \url{https://arxiv.org/abs/2503.14492}.

\bibitem[Zhu et~al.(2025{\natexlab{a}})Zhu, Zou, Jiang, Sun, Casser, Huang,
  Wang, Yang, Gao, Guibas, Tan, and Anguelov]{scenecrafter2025}
Zehao Zhu, Yuliang Zou, Chiyu~Max Jiang, Bo~Sun, Vincent Casser, Xiukun Huang,
  Jiahao Wang, Zhenpei Yang, Ruiqi Gao, Leonidas Guibas, Mingxing Tan, and
  Dragomir Anguelov.
\newblock Scenecrafter: Controllable multi-view driving scene editing.
\newblock In \emph{Proceedings of the IEEE/CVF Conference on Computer Vision
  and Pattern Recognition}, pages 6812--6822, 2025{\natexlab{a}}.
\newblock \doi{10.1109/CVPR52734.2025.00639}.

\bibitem[Lin et~al.(2025)Lin, Wang, Liang, Zhang, Fidler, Wang, and
  Gojcic]{lin2025weatherweaver}
Chih-Hao Lin, Zian Wang, Ruofan Liang, Yuxuan Zhang, Sanja Fidler, Shenlong
  Wang, and Zan Gojcic.
\newblock Controllable weather synthesis and removal with video diffusion
  models.
\newblock In \emph{Proceedings of the IEEE/CVF International Conference on
  Computer Vision}, pages 13580--13591, 2025.
\newblock \doi{10.1109/ICCV51701.2025.01261}.

\bibitem[Yan et~al.(2024)Yan, Lin, Zhou, Wang, Sun, Zhan, Lang, Zhou, and
  Peng]{yan2024streetgaussians}
Yunzhi Yan, Haotong Lin, Chenxu Zhou, Weijie Wang, Haiyang Sun, Kun Zhan,
  Xianpeng Lang, Xiaowei Zhou, and Sida Peng.
\newblock Street gaussians: Modeling dynamic urban scenes with gaussian
  splatting.
\newblock In \emph{European Conference on Computer Vision}, pages 156--173.
  Springer, 2024.
\newblock \doi{10.1007/978-3-031-73464-9_10}.

\bibitem[Zhou et~al.(2024{\natexlab{a}})Zhou, Lin, Shan, Wang, Sun, and
  Yang]{zhou2024drivinggaussian}
Xiaoyu Zhou, Zhiwei Lin, Xiaojun Shan, Yongtao Wang, Deqing Sun, and Ming-Hsuan
  Yang.
\newblock Drivinggaussian: Composite gaussian splatting for surrounding dynamic
  autonomous driving scenes.
\newblock In \emph{Proceedings of the IEEE/CVF Conference on Computer Vision
  and Pattern Recognition}, pages 21634--21643, 2024{\natexlab{a}}.
\newblock \doi{10.1109/CVPR52733.2024.02044}.

\bibitem[Zhou et~al.(2024{\natexlab{b}})Zhou, Lin, Wang, Lu, Bai, Liu, Wang,
  Geiger, and Liao]{zhou2024hugsim}
Hongyu Zhou, Longzhong Lin, Jiabao Wang, Yichong Lu, Dongfeng Bai, Bingbing
  Liu, Yue Wang, Andreas Geiger, and Yiyi Liao.
\newblock {HUGSIM}: A real-time, photo-realistic and closed-loop simulator for
  autonomous driving, 2024{\natexlab{b}}.
\newblock URL \url{https://arxiv.org/abs/2412.01718}.

\bibitem[Hess et~al.(2025)Hess, Lindstr{\"o}m, Fatemi, Petersson, and
  Svensson]{splatad2024}
Georg Hess, Carl Lindstr{\"o}m, Maryam Fatemi, Christoffer Petersson, and
  Lennart Svensson.
\newblock {SplatAD}: Real-time lidar and camera rendering with {3D} gaussian
  splatting for autonomous driving.
\newblock In \emph{Proceedings of the IEEE/CVF Conference on Computer Vision
  and Pattern Recognition}, 2025.
\newblock \doi{10.1109/CVPR52734.2025.01119}.

\bibitem[Wang et~al.(2026)Wang, Pittaluga, Tasneem, You, Chandraker, and
  Jiang]{horizonforge2026}
Yifan Wang, Francesco Pittaluga, Zaid Tasneem, Chenyu You, Manmohan Chandraker,
  and Ziyu Jiang.
\newblock {HorizonForge}: Driving scene editing with any trajectories and any
  vehicles, 2026.
\newblock URL \url{https://arxiv.org/abs/2602.21333}.

\bibitem[Xu et~al.(2025{\natexlab{b}})Xu, Wang, Zhao, Fu, and
  Gao]{streetunveiler2025}
Jingwei Xu, Yikai Wang, Yiqun Zhao, Yanwei Fu, and Shenghua Gao.
\newblock {3D} streetunveiler with semantic-aware {2DGS}: A simple baseline.
\newblock In \emph{International Conference on Learning Representations},
  2025{\natexlab{b}}.
\newblock URL \url{https://openreview.net/forum?id=G6aJyS0ZV0}.
\newblock arXiv:2405.18416.

\bibitem[Mao et~al.(2025)Mao, Li, Ivanovic, Chen, Wang, You, Xiao, Xu, Pavone,
  and Wang]{dreamdrive2025}
Jiageng Mao, Boyi Li, Boris Ivanovic, Yuxiao Chen, Yan Wang, Yurong You,
  Chaowei Xiao, Danfei Xu, Marco Pavone, and Yue Wang.
\newblock {DreamDrive}: Generative {4D} scene modeling from street view images,
  2025.
\newblock URL \url{https://arxiv.org/abs/2501.00601}.

\bibitem[Zhao et~al.(2024)Zhao, Ni, Wang, Zhu, Zhang, Wang, Huang, Chen, Wang,
  Zhang, Mei, and Wang]{drivedreamer4d2024}
Guosheng Zhao, Chaojun Ni, Xiaofeng Wang, Zheng Zhu, Xueyang Zhang, Yida Wang,
  Guan Huang, Xinze Chen, Boyuan Wang, Youyi Zhang, Wenjun Mei, and Xingang
  Wang.
\newblock {DriveDreamer4D}: World models are effective data machines for {4D}
  driving scene representation, 2024.
\newblock URL \url{https://arxiv.org/abs/2410.13571}.

\bibitem[Zhu et~al.(2025{\natexlab{b}})Zhu, Wu, Zhu, Zhou, Sun, Wan, Ma, Chen,
  Ye, Xie, and Yang]{worldsplat2025}
Ziyue Zhu, Zhanqian Wu, Zhenxin Zhu, Lijun Zhou, Haiyang Sun, Bing Wan, Kun Ma,
  Guang Chen, Hangjun Ye, Jin Xie, and Jian Yang.
\newblock {WorldSplat}: Gaussian-centric feed-forward {4D} scene generation for
  autonomous driving, 2025{\natexlab{b}}.
\newblock URL \url{https://arxiv.org/abs/2509.23402}.

\bibitem[Liu et~al.(2026)Liu, Xiong, Luo, Zhang, Li, Liu, and
  Tan]{autoweather4d2026}
Tianyu Liu, Weitao Xiong, Kunming Luo, Manyuan Zhang, Peng Li, Yuan Liu, and
  Ping Tan.
\newblock {AutoWeather4D}: Autonomous driving video weather conversion via
  {G}-buffer dual-pass editing, 2026.
\newblock URL \url{https://arxiv.org/abs/2603.26546}.

\bibitem[Xie et~al.(2024)Xie, Zong, Qiu, Li, Feng, Yang, and
  Jiang]{physgaussian2024}
Tianyi Xie, Zeshun Zong, Yuxing Qiu, Xuan Li, Yutao Feng, Yin Yang, and
  Chenfanfu Jiang.
\newblock Physgaussian: Physics-integrated {3D} gaussians for generative
  dynamics.
\newblock In \emph{Proceedings of the IEEE/CVF Conference on Computer Vision
  and Pattern Recognition}, pages 4389--4398, 2024.
\newblock \doi{10.1109/CVPR52733.2024.00420}.

\bibitem[Qiu et~al.(2024)Qiu, Yang, Zeng, and Wang]{featuresplatting2024}
Ri-Zhao Qiu, Ge~Yang, Weijia Zeng, and Xiaolong Wang.
\newblock Feature splatting: Language-driven physics-based scene synthesis and
  editing.
\newblock In \emph{European Conference on Computer Vision}, 2024.
\newblock \doi{10.1007/978-3-031-72940-9_21}.
\newblock arXiv:2404.01223.

\bibitem[Huang et~al.(2025)Huang, Zhang, Zeng, Zhang, Li, Zuo, and
  Lau]{dreamphysics2025}
Tianyu Huang, Haoze Zhang, Yihan Zeng, Zhilu Zhang, Hui Li, Wangmeng Zuo, and
  Rynson W.~H. Lau.
\newblock Dreamphysics: Learning physics-based {3D} dynamics with video
  diffusion priors.
\newblock In \emph{Proceedings of the AAAI Conference on Artificial
  Intelligence}, 2025.
\newblock \doi{10.1609/aaai.v39i4.32389}.
\newblock arXiv:2406.01476.

\bibitem[Liu et~al.(2024{\natexlab{a}})Liu, Wang, Yao, Zhang, Zhou, and
  Duan]{physics3d2024}
Fangfu Liu, Hanyang Wang, Shunyu Yao, Shengjun Zhang, Jie Zhou, and Yueqi Duan.
\newblock Physics3d: Learning physical properties of {3D} gaussians via video
  diffusion, 2024{\natexlab{a}}.
\newblock URL \url{https://arxiv.org/abs/2406.04338}.

\bibitem[Zhao et~al.(2025{\natexlab{b}})Zhao, Wang, Zhao, Fei, Wang, Long, and
  Zou]{physsplat2025}
Haoyu Zhao, Hao Wang, Xingyue Zhao, Hao Fei, Hongqiu Wang, Chengjiang Long, and
  Hua Zou.
\newblock Efficient physics simulation for {3D} scenes via {MLLM}-guided
  gaussian splatting.
\newblock In \emph{Proceedings of the IEEE/CVF International Conference on
  Computer Vision}, 2025{\natexlab{b}}.
\newblock \doi{10.1109/ICCV51701.2025.00498}.
\newblock arXiv:2411.12789.

\bibitem[Li et~al.(2026)Li, Shen, Ni, Pan, Zhang, and Zhu]{ngff2026}
Shiqian Li, Ruihong Shen, Junfeng Ni, Chang Pan, Chi Zhang, and Yixin Zhu.
\newblock Learning physics-grounded {4D} dynamics with neural gaussian force
  fields, 2026.
\newblock URL \url{https://arxiv.org/abs/2602.00148}.

\bibitem[Huang et~al.(2026{\natexlab{a}})Huang, Azfar, Shi, and
  Ke]{huang2026real2sim}
Kaicong Huang, Talha Azfar, Weisong Shi, and Ruimin Ke.
\newblock Real2sim: A physics-driven and editable gaussian splatting framework
  for autonomous driving scenes, 2026{\natexlab{a}}.
\newblock URL \url{https://arxiv.org/abs/2605.13591}.

\bibitem[Huang et~al.(2026{\natexlab{b}})Huang, Ma, and Ke]{huang2026carlags}
Kaicong Huang, Meng Ma, and Ruimin Ke.
\newblock {CARLA-GS}: Decoupling representation, reasoning, and physics
  simulation for autonomous driving corner-case synthesis, 2026{\natexlab{b}}.
\newblock URL \url{https://arxiv.org/abs/2607.07601}.

\bibitem[Zhou et~al.(2026)Zhou, Zhang, Sun, Wang, Wen, Kong, Wu, and
  Ma]{physicsawaregs2026}
Feng Zhou, Jian Zhang, Yuhang Sun, He~Wang, Qiong Wen, Debao Kong, Tieru Wu,
  and Rui Ma.
\newblock Physics-aware {3D} gaussian editing for driving scene generation,
  2026.
\newblock URL \url{https://arxiv.org/abs/2605.25373}.

\bibitem[Liu et~al.(2024{\natexlab{b}})Liu, Ren, Gupta, and Wang]{physgen2024}
Shaowei Liu, Zhongzheng Ren, Saurabh Gupta, and Shenlong Wang.
\newblock Physgen: Rigid-body physics-grounded image-to-video generation.
\newblock In \emph{European Conference on Computer Vision}, 2024{\natexlab{b}}.
\newblock URL \url{https://arxiv.org/abs/2409.18964}.
\newblock arXiv:2409.18964.

\bibitem[Yang et~al.(2025{\natexlab{a}})Yang, Li, Zhang, Yin, Bai, Ma, Wang,
  Cai, Wong, Lu, and Jia]{vlipp2025}
Xindi Yang, Baolu Li, Yiming Zhang, Zhenfei Yin, Lei Bai, Liqian Ma, Zhiyong
  Wang, Jianfei Cai, Tien-Tsin Wong, Huchuan Lu, and Xu~Jia.
\newblock {VLIPP}: Towards physically plausible video generation with vision
  and language informed physical prior.
\newblock In \emph{Proceedings of the IEEE/CVF International Conference on
  Computer Vision}, 2025{\natexlab{a}}.
\newblock URL \url{https://arxiv.org/abs/2503.23368}.
\newblock arXiv:2503.23368.

\bibitem[Wang et~al.(2025{\natexlab{a}})Wang, Ma, Cao, Zheng, Zhang, Feng, Liu,
  Ma, Cheng, Leng, Yin, and Liang]{wisa2025}
Jing Wang, Ao~Ma, Ke~Cao, Jun Zheng, Zhanjie Zhang, Jiasong Feng, Shanyuan Liu,
  Yuhang Ma, Bo~Cheng, Dawei Leng, Yuhui Yin, and Xiaodan Liang.
\newblock {WISA}: World simulator assistant for physics-aware text-to-video
  generation, 2025{\natexlab{a}}.
\newblock URL \url{https://arxiv.org/abs/2503.08153}.

\bibitem[Pathak et~al.(2026)Pathak, Arani, Pechenizkiy, and
  Zonooz]{pathak2026physvid}
Saurabh Pathak, Elahe Arani, Mykola Pechenizkiy, and Bahram Zonooz.
\newblock Physvid: Physics-aware local conditioning for generative video
  models.
\newblock In \emph{Proceedings of the IEEE/CVF Conference on Computer Vision
  and Pattern Recognition}, 2026.
\newblock URL \url{https://arxiv.org/abs/2603.26285}.
\newblock arXiv:2603.26285.

\bibitem[Yang et~al.(2025{\natexlab{b}})Yang, Liu, Lu, Hou, Miao, Peng, Feng,
  Bai, and Zhao]{geniedrive2025}
Zhenya Yang, Zhe Liu, Yuxiang Lu, Liping Hou, Chenxuan Miao, Siyi Peng, Bailan
  Feng, Xiang Bai, and Hengshuang Zhao.
\newblock {GenieDrive}: Towards physics-aware driving world model with {4D}
  occupancy guided video generation, 2025{\natexlab{b}}.
\newblock URL \url{https://arxiv.org/abs/2512.12751}.

\bibitem[Wong et~al.(2026)Wong, Jia, You, Zhang, Xu, Xia, Qiu, Zhang, Zhao,
  Yan, Zhou, Chen, Guo, Xu, and Yan]{pointasskeleton2026}
Songbur Wong, Xiaosong Jia, Junqi You, Bo~Zhang, Pei Xu, Renqiu Xia, Yuping
  Qiu, Shaofeng Zhang, Zelin Zhao, Xuechao Yan, Yuchen Zhou, Yurui Chen, Wen
  Guo, Hang Xu, and Junchi Yan.
\newblock Point as skeleton: Accumulated point cloud enhanced autoregressive
  generation for closed-loop autonomous driving simulation, 2026.
\newblock URL \url{https://arxiv.org/abs/2607.06516}.

\bibitem[Zhu et~al.(2026)Zhu, Chen, Li, and Bian]{scenefactory2026}
Yicheng Zhu, Yang Chen, Tao Li, and Zilin Bian.
\newblock Scenefactory: {GPU}-accelerated multi-agent driving simulation with
  physics-based vehicle dynamics, 2026.
\newblock URL \url{https://arxiv.org/abs/2605.08528}.

\bibitem[Dosovitskiy et~al.(2017)Dosovitskiy, Ros, Codevilla, Lopez, and
  Koltun]{carla2017}
Alexey Dosovitskiy, German Ros, Felipe Codevilla, Antonio Lopez, and Vladlen
  Koltun.
\newblock {CARLA}: An open urban driving simulator.
\newblock In \emph{Proceedings of the 1st Annual Conference on Robot Learning},
  volume~78 of \emph{Proceedings of Machine Learning Research}, pages 1--16.
  PMLR, 2017.
\newblock URL \url{https://proceedings.mlr.press/v78/dosovitskiy17a.html}.

\bibitem[{Mechanical Simulation Corporation}(2026)]{carsim}
{Mechanical Simulation Corporation}.
\newblock {CarSim}: Vehicle dynamics simulation software.
\newblock \url{https://www.carsim.com/}, 2026.
\newblock Accessed: 2026-07-01.

\bibitem[Sun et~al.(2020)Sun, Kretzschmar, Dotiwalla, Chouard, Patnaik, Tsui,
  Guo, Zhou, Chai, Caine, Vasudevan, Han, Ngiam, Zhao, Timofeev, Ettinger,
  Krivokon, Gao, Joshi, Zhang, Shlens, Chen, and Anguelov]{sun2020waymo}
Pei Sun, Henrik Kretzschmar, Xerxes Dotiwalla, Aur{\'e}lien Chouard, Vijaysai
  Patnaik, Paul Tsui, James Guo, Yin Zhou, Yuning Chai, Benjamin Caine, Vijay
  Vasudevan, Wei Han, Jiquan Ngiam, Hang Zhao, Aleksei Timofeev, Scott
  Ettinger, Maxim Krivokon, Amy Gao, Aditya Joshi, Yu~Zhang, Jon Shlens,
  Zhifeng Chen, and Dragomir Anguelov.
\newblock Scalability in perception for autonomous driving: Waymo open dataset.
\newblock In \emph{Proceedings of the IEEE/CVF Conference on Computer Vision
  and Pattern Recognition}, pages 2446--2454, 2020.

\bibitem[Geyer et~al.(2020)Geyer, Kassahun, Mahmudi, Ricou, Durgesh, Chung,
  Hauswald, Pham, Muehlegg, Dorn, Fernandez, Jaenicke, Mirashi, Savani, Sturm,
  Vorobiov, and Oelker]{geyer2020a2d2}
Jakob Geyer, Yohannes Kassahun, Mentar Mahmudi, Xavier Ricou, Rupesh Durgesh,
  Andrew~S. Chung, Lorenz Hauswald, Viet~Hoang Pham, Maximilian Muehlegg,
  Sebastian Dorn, Tiffany Fernandez, Martin Jaenicke, Sudesh Mirashi,
  Chiragkumar Savani, Martin Sturm, Oleksandr Vorobiov, and Martin Oelker.
\newblock {A2D2}: Audi autonomous driving dataset, 2020.

\bibitem[Wang et~al.(2025{\natexlab{b}})Wang, Chen, Karaev, Vedaldi, Rupprecht,
  and Novotny]{wang2025vggt}
Jianyuan Wang, Minghao Chen, Nikita Karaev, Andrea Vedaldi, Christian
  Rupprecht, and David Novotny.
\newblock Vggt: Visual geometry grounded transformer.
\newblock In \emph{Proceedings of the IEEE/CVF Conference on Computer Vision
  and Pattern Recognition}, 2025{\natexlab{b}}.

\bibitem[Wu et~al.(2023)Wu, Zhang, Liao, Chen, Hou, Wang, Sun, Yan, and
  Lin]{wu2023dover}
Haoning Wu, Erli Zhang, Liang Liao, Chaofeng Chen, Jingwen Hou, Annan Wang,
  Wenxiu Sun, Qiong Yan, and Weisi Lin.
\newblock Exploring video quality assessment on user generated contents from
  aesthetic and technical perspectives.
\newblock In \emph{Proceedings of the IEEE/CVF International Conference on
  Computer Vision}, pages 7591--7601, 2023.
\newblock URL \url{https://arxiv.org/abs/2211.04894}.

\bibitem[Radford et~al.(2021)Radford, Kim, Hallacy, Ramesh, Goh, Agarwal,
  Sastry, Askell, Mishkin, Clark, Krueger, and Sutskever]{radford2021clip}
Alec Radford, Jong~Wook Kim, Chris Hallacy, Aditya Ramesh, Gabriel Goh,
  Sandhini Agarwal, Girish Sastry, Amanda Askell, Pamela Mishkin, Jack Clark,
  Gretchen Krueger, and Ilya Sutskever.
\newblock Learning transferable visual models from natural language
  supervision.
\newblock In \emph{Proceedings of the 38th International Conference on Machine
  Learning}, volume 139 of \emph{Proceedings of Machine Learning Research},
  pages 8748--8763. PMLR, 2021.
\newblock URL \url{https://proceedings.mlr.press/v139/radford21a.html}.

\bibitem[HaCohen et~al.(2024)HaCohen, Chiprut, Brazowski, Shalem, Moshe,
  Richardson, Levin, Shiran, Zabari, Gordon, Panet, Weissbuch, Kulikov,
  Bitterman, Melumian, and Bibi]{HaCohen2024LTXVideo}
Yoav HaCohen, Nisan Chiprut, Benny Brazowski, Daniel Shalem, Dudu Moshe, Eitan
  Richardson, Eran Levin, Guy Shiran, Nir Zabari, Ori Gordon, Poriya Panet,
  Sapir Weissbuch, Victor Kulikov, Yaki Bitterman, Zeev Melumian, and Ofir
  Bibi.
\newblock Ltx-video: Realtime video latent diffusion, 2024.

\bibitem[Burckhardt(1993)]{burckhardt1993}
Manfred Burckhardt.
\newblock \emph{Fahrwerktechnik: Radschlupf-Regelsysteme}.
\newblock Vogel, W{\"u}rzburg, Germany, 1993.

\bibitem[Wu et~al.(2022)Wu, Chen, Hou, Liao, Wang, Sun, Yan, and
  Lin]{wu2022fastvqa}
Haoning Wu, Chaofeng Chen, Jingwen Hou, Liang Liao, Annan Wang, Wenxiu Sun,
  Qiong Yan, and Weisi Lin.
\newblock {FAST-VQA}: Efficient end-to-end video quality assessment with
  fragment sampling.
\newblock In \emph{European Conference on Computer Vision}, pages 538--554.
  Springer, 2022.
\newblock URL \url{https://arxiv.org/abs/2207.02595}.

\bibitem[Zhang et~al.(2018)Zhang, Isola, Efros, Shechtman, and
  Wang]{zhang2018lpips}
Richard Zhang, Phillip Isola, Alexei~A. Efros, Eli Shechtman, and Oliver Wang.
\newblock The unreasonable effectiveness of deep features as a perceptual
  metric.
\newblock In \emph{Proceedings of the IEEE Conference on Computer Vision and
  Pattern Recognition}, pages 586--595, 2018.
\newblock URL \url{https://arxiv.org/abs/1801.03924}.

\bibitem[Gallaway et~al.(1971)Gallaway, Schiller, and
  Rose]{gallaway1971waterdepth}
B.~M. Gallaway, R.~E. Schiller, Jr., and J.~G. Rose.
\newblock The effects of rainfall intensity, pavement cross slope, surface
  texture, and drainage length on pavement water depths.
\newblock Technical Report Research Report 138-5, Texas Transportation
  Institute, Texas A\&M University, May 1971.
\newblock URL
  \url{https://static.tti.tamu.edu/tti.tamu.edu/documents/138-5.pdf}.

\bibitem[Gallaway et~al.(1979)Gallaway, Ivey, Hayes, Ledbetter, Olson, Woods,
  and Schiller]{gallaway1979hydroplaning}
B.~M. Gallaway, D.~L. Ivey, G.~G. Hayes, W.~B. Ledbetter, R.~M. Olson, D.~L.
  Woods, and R.~F. Schiller.
\newblock Pavement and geometric design criteria for minimizing hydroplaning.
\newblock Technical Report FHWA-RD-79-31, Federal Highway Administration,
  December 1979.
\newblock URL \url{https://trid.trb.org/view/145115}.

\bibitem[Wallman and {\AA}str{\"o}m(2001)]{wallman2001roadfriction}
Carl-Gustaf Wallman and Henrik {\AA}str{\"o}m.
\newblock Friction measurement methods and the correlation between road
  friction and traffic safety: A literature review.
\newblock Technical Report VTI meddelande 911A, Swedish National Road and
  Transport Research Institute (VTI), Link{\"o}ping, Sweden, 2001.
\newblock URL
  \url{https://vti.diva-portal.org/smash/get/diva2:673366/FULLTEXT01.pdf}.

\bibitem[Ivan et~al.(2010)Ivan, Ravishanker, Jackson, Aronov, and
  Guo]{ivan2010wetfriction}
John~N. Ivan, Nalini Ravishanker, Eric Jackson, Brien Aronov, and Sizhen Guo.
\newblock Incorporating wet pavement friction into traffic safety analysis.
\newblock Technical Report JHR 10-324, Connecticut Transportation Institute,
  University of Connecticut, November 2010.
\newblock URL \url{https://rosap.ntl.bts.gov/view/dot/48821/dot_48821_DS1.pdf}.

\bibitem[Boz et~al.(2023)Boz, Flintsch, and
  de~Le\'{o}n~Izeppi]{boz2023densegraded}
Ilker Boz, Gerardo~W. Flintsch, and Edgar de~Le\'{o}n~Izeppi.
\newblock Functional characteristics of dense-graded asphalt surface mixtures.
\newblock Technical Report FHWA/VTRC 23-R15, Virginia Transportation Research
  Council, April 2023.
\newblock URL
  \url{https://vtrc.virginia.gov/media/vtrc/vtrc-pdf/vtrc-pdf/23-R15.pdf}.

\bibitem[{ASTM International}(2019)]{astmE965}
{ASTM International}.
\newblock Astm e965/e965m-15(2019): Standard test method for measuring pavement
  macrotexture depth using a volumetric technique.
\newblock Technical report, ASTM International, West Conshohocken, PA, 2019.
\newblock URL \url{https://www.astm.org/e0965-15r19.html}.

\bibitem[Kilgore et~al.(2024)Kilgore, Atayee, and Herrmann]{kilgore2024hec22}
R.~Kilgore, A.~T. Atayee, and G.~R. Herrmann.
\newblock Urban drainage design.
\newblock Technical Report Hydraulic Engineering Circular No. 22, Fourth
  Edition, FHWA-HIF-24-006, Federal Highway Administration, February 2024.
\newblock URL
  \url{https://www.fhwa.dot.gov/engineering/hydraulics/pubs/hif24006.pdf}.

\bibitem[Judson and Doesken(2000)]{judson2000freshsnowdensity}
Arthur Judson and Nolan Doesken.
\newblock Density of freshly fallen snow in the central rocky mountains.
\newblock \emph{Bulletin of the American Meteorological Society}, 81\penalty0
  (7):\penalty0 1577--1587, 2000.
\newblock \doi{10.1175/1520-0477(2000)081<1577:DOFFSI>2.3.CO;2}.
\newblock URL \url{https://climate.colostate.edu/pdfs/SnowDensity_BAMS.pdf}.

\bibitem[Ichihara and Mizoguchi(1970)]{ichihara1970snowfriction}
Kaoru Ichihara and Mitsumasa Mizoguchi.
\newblock Skid resistance of snow- or ice-covered roads.
\newblock In \emph{Snow Removal and Ice Control Research}, number 115 in
  Highway Research Board Special Report, pages 104--114. Highway Research
  Board, 1970.
\newblock URL \url{https://onlinepubs.trb.org/Onlinepubs/sr/sr115/115-010.pdf}.

\end{thebibliography}
\endgroup

\clearpage
\appendix
\suppressfloats[t]
\raggedbottom
\renewcommand{\thefigure}{S\arabic{figure}}
\renewcommand{\thetable}{S\arabic{table}}
\renewcommand{\theequation}{S\arabic{equation}}
\renewcommand{\theHfigure}{appendix.S\arabic{figure}}
\renewcommand{\theHtable}{appendix.S\arabic{table}}
\renewcommand{\theHequation}{appendix.S\arabic{equation}}
\setcounter{figure}{0}
\setcounter{table}{0}
\setcounter{equation}{0}
\setcounter{topnumber}{4}
\setcounter{bottomnumber}{2}
\setcounter{totalnumber}{6}
\setcounter{dbltopnumber}{3}
\renewcommand{\topfraction}{0.95}
\renewcommand{\bottomfraction}{0.90}
\renewcommand{\textfraction}{0.05}
\renewcommand{\floatpagefraction}{0.85}
\renewcommand{\dbltopfraction}{0.95}
\renewcommand{\dblfloatpagefraction}{0.85}

Appendix~A defines the evaluation protocols, Appendix~B provides the complete
vehicle-state update, Appendix~C reports the CarSim and A2D2 evaluations,
Appendix~D presents additional video and visual results, and Appendix~E reports
the bounded surface-parameter sensitivity.

\section{Evaluation Protocols and Reproducibility}
\label{app:protocols}

\subsection{Experiment map}

Table~\ref{tab:experiment_map_supp} separates the 19-case calibration
set from the frozen four-case speed check and 12-case controlled holdout suite. Video and
visual evaluations do not modify the vehicle model.

\begin{table*}[t]
\centering
\small
\setlength{\tabcolsep}{4.5pt}
\begin{tabular}{p{0.16\textwidth}p{0.25\textwidth}p{0.34\textwidth}p{0.14\textwidth}}
\toprule
Component & Evaluation set & Primary comparison & Parameter status \\
\midrule
Vehicle-model calibration & 19 CarSim development cases: 9 flat braking,
5 speed humps, and 5 road depressions & Original-time speed, body pitch, slip, normal
load, and event outcomes & Calibrated on development set \\
Initial-speed check & 4 road-feature cases at unseen initial speeds & Same
CarSim state and event metrics & Frozen \\
Frozen controlled holdout evaluation & 12 cases: 5 within-family weather-severity,
3 brake-command, and 4
road-elevation-profile holdouts &
CarSim state RMSE and stopping-distance or feature-center error & Frozen \\
Cross-drive real-vehicle brake-dive check after vehicle-specific calibration &
14 retained braking events from 3 A2D2 Audi Q7
drives & Brake-pressure-driven pitch angle and pitch rate over the first
$0.5\,$s & Leave one complete drive out \\
Video-response protocols & 8 non-sunny weather conditions, 2 geometric-road-hazard
tasks, 2 deterministic geometry-only CarSim-pose video-reference clips,
and 9 motion-preservation clips; generator results use 3 seeds &
VGGT-recovered video response against
CarSim, input--output preservation, and an empirical rendering/VGGT recovery
reference, plus lateral path deviation & Frozen \\
Visual control & 9 weather conditions (5 displayed) and paired
speed-hump/road-depression edits &
Matched-frame appearance, quality/severity, localized edit, road-anchor, and
temporal-mask metrics & Frozen \\
\bottomrule
\end{tabular}
\caption{Map of the experiment sets. ``Frozen'' means that no calibration
parameter is updated using the listed evaluation.}
\label{tab:experiment_map_supp}
\end{table*}

\subsection{CarSim signal comparison}
\label{app:carsim_protocol}

CarSim signals are evaluated on their original time axis, including the static
body-pitch offset. muSync-GS outputs are sampled at the corresponding CarSim
timestamps. All predefined cases are retained, without post hoc temporal
alignment, pitch-offset removal, or residual correction. Slip is scored above
$0.5\,\mathrm{m/s}$, and wheel-level slip and load RMSEs are macro-averaged.

For flat-road braking, the event window begins at brake application and ends
one second after the CarSim stop. For road-elevation-profile cases, it begins
$0.5\,\mathrm{m}$ before front-axle feature entry and ends $0.5\,\mathrm{m}$
after rear-axle exit. The event metric is stopping-distance absolute error for
flat-road cases and feature-center speed absolute error for road-profile cases.

\subsection{Video-baseline conditioning}
\label{app:baseline_conditioning}

We compare Cosmos-Transfer2.5-2B~\cite{cosmostransfer2025} (hereafter Cosmos),
LTX-Video~\cite{HaCohen2024LTXVideo}, and VACE~\cite{vace2025}. The three
baseline generators receive the same positive task prompt within each
evaluation protocol, while auxiliary conditioning and negative-prompt handling
follow each model's native interface. We also match the source clip, output
length, and visual seeds 2025--2027.
Table~\ref{tab:baseline_conditioning_supp} lists the model versions, principal
inputs, and frozen inference settings used for the submitted results.
Representative prompt instances are shown below, while all condition-specific
prompts and commands are retained in the frozen run manifests.

\pagebreak[4]
For weather response, the shared positive prompt specifies the target weather
condition and precipitation level, without requesting preservation of the
source trajectory or providing an explicit vehicle- or camera-motion target.
For geometry response, the shared positive prompt specifies the road-feature
type, supplied road location, and geometric dimensions, without prescribing a
source trajectory, camera pitch, heave, response amplitude, or response timing.
Motion preservation uses a separate prompt that explicitly requests retention
of the supplied braking and suspension motion. No response prompt contains a
CarSim-derived signal or target camera pose.

\paragraph{Prompt instances.}
Representative Rain-50 and video-speed-bump prompts are shown below. Other
weather rates and flat/depression variants use the corresponding
condition-specific wording recorded in the run manifests.
In the surrounding discussion, \emph{video speed bump} denotes the short
$0.60\,\mathrm{m}$ prompt feature, whereas \emph{CarSim speed hump} denotes the
$2.5$--$3.7\,\mathrm{m}$ road profiles.
{\small

\noindent\textbf{Weather response (all three baselines, Rain-50).}
``A realistic forward-facing dashcam video in very heavy rain at
50 millimeters per hour, with dense visible rainfall, wet reflective asphalt,
strong water spray, puddles, and realistic rain haze.''

\noindent\textbf{Geometry response (all three baselines, video speed bump).}
``Insert a single realistic asphalt speed bump across the ego driving lane at
the supplied road location. The speed bump is 0.60 meters long in the travel
direction and 0.10 meters high, with realistic road contact, perspective,
texture, lighting, shadows, and temporal consistency as the vehicle approaches
and passes the feature.''

\noindent\textbf{Motion preservation (all three baselines, Rain-50 flat).}
``A realistic forward-facing dashcam video of the same passenger vehicle on the
same suburban residential street. Faithfully preserve the supplied scene
geometry, camera trajectory, braking and suspension motion. The weather is very
heavy rain at 50 millimeters per hour, dense visible rainfall, wet reflective
asphalt, strong water spray, puddles and realistic rain haze. The asphalt road
remains completely flat. Do not add a speed hump, depression, pothole, trench,
ramp, or any other road obstacle.''
}

Cosmos and LTX use $480\times320$ RGB streams; Cosmos additionally receives
source depth. VACE's 480p interface produces
$768\times512$ controls and outputs, which are canonicalized to H.264 at
$480\times320$ for evaluation. Every evaluated sequence contains 81 frames.
Weather response and motion preservation use 10 fps. VGGT geometry-response
metrics use the native 16-fps output timestamps and the fixed
$2.2$--$3.8\,$s clip-time interval, with CarSim sampled on the same grid. The
16-to-10-fps retiming described below is not used for this physical-time
response metric. No score uses content-dependent frame selection or a
method-specific temporal shift.

\subsection{muSync-GS rendering implementation}
\label{app:renderer_implementation}

WeatherEdit~\cite{weatheredit2025} supplies the static atmospheric transform
and Gaussian-particle renders. For every ego pose, paired renders with and
without particles define the signed display-space residual
\begin{equation}
\Delta I_{\mathrm{part}}(t)=
I_{\mathrm{with\mbox{-}part}}(t)-I_{\mathrm{no\mbox{-}part}}(t),
\end{equation}
which is added after pose synchronization. Particle count is a monotonic
function of precipitation rate, while road accumulation is controlled
separately by the surface state.

Rain uses semantic-road, RGB, depth, and normal buffers with reflection,
refraction, and grazing-angle wet-road compositing following the organization
of RainyGS~\cite{dai2025rainygs}. Exact visual anchors are the WeatherEdit
wet-road baseline at $2\,\mathrm{mm/h}$, wet sheen at $10\,\mathrm{mm/h}$,
and progressively stronger static thin-film composites at 25 and
$50\,\mathrm{mm/h}$. An intermediate rate linearly blends the adjacent
pose-matched static anchors; its filtered particle residual uses the analogous
weight from the continuous target particle count. Dynamic shallow-water
ripples, impacts, and splashes are disabled in the reported videos.

Snow adapts the normal-guided accumulation principle of
Weather-Magician~\cite{sang2025weathermagician}. A Gaussian is retained as a
road anchor only if it is registered to the reconstructed semantic road, lies
within the stored signed-height tolerance, and satisfies
$\mathbf n_i^\top\mathbf u\geq\cos30^\circ$. Normals use the shortest Gaussian
axis when reliable and otherwise local PCA. A seed-conditioned deterministic coarse-to-fine
field $\xi_i\in[0,1]$ activates anchor $i$ when $\xi_i\leq c_s$, producing
nested support as coverage increases. Local neighbor planes are
footprint-filtered and offset by target snow depth plus micro-height variation.
The same seed and scene assets reproduce the same coverage field.

\subsection{Video-motion protocols and metrics}
\label{app:video_protocol}

All weather methods use visual seeds 2025--2027. In muSync-GS, only particles
and spatial snow realization vary; precipitation/depth controls, friction,
telemetry, station, and camera poses remain bitwise frozen. Road geometry
is deterministic (\emph{det.}). Every clip uses the same camera-only VGGT
procedure~\cite{wang2025vggt}. Three protocols separate response generation
from preservation:
\begin{itemize}
    \item \textbf{Weather-response diagnostic:} the three video baselines
    receive the same sunny braking source and the same positive prompt
    specifying the target weather condition and precipitation level, without
    any trajectory-preservation instruction or explicit motion target.
    muSync-GS is rendered under the corresponding numerical precipitation
    control. For every method, the recovered normalized braking progress is
    compared with the matched normalized CarSim response.

    \item \textbf{Geometry-response diagnostic:} the three video baselines
    receive the same sunny flat-road source and the same positive prompt
    specifying the road-feature type, supplied location, and geometric
    dimensions, without a target camera-motion response. muSync-GS is rendered
    using the corresponding road-elevation input. For every method, the
    recovered event-local camera pitch is compared with the matched CarSim
    reference.

    \item \textbf{Motion-preservation protocol:} this baseline-only protocol
    uses an input containing condition-specific motion and a separate prompt
    that explicitly requests retention of the supplied braking or suspension
    response.
\end{itemize}
For each seed, the motion-preservation set contains all nine combinations of
three weather conditions (sunny, Rain-50, and Snow-3) and three motions
(braking, bump, and depression). Only its six road-event clips enter the
event-local pitch-correlation
diagnostic; the three flat-road braking clips remain part of the preservation
set but do not contain a localized road-pitch event.

Because monocular translation has unknown scale, forward progress is normalized
independently for each sequence. Let $\widehat p_c(t)$ and $\widehat p_0(t)$
denote VGGT-recovered normalized progress for condition $c$ and its sunny
control, and let $p_c^{\mathrm{CS}}(t)$ and $p_0^{\mathrm{CS}}(t)$ denote the
matched CarSim quantities. On the fixed $T=8\,$s window, raw recovered forward
coordinate is resampled to the common 81-frame grid and smoothed before
terminal-direction sign correction and min--max normalization to $[0,1]$.
Normalized progress is smoothed once more, monotonized by a cumulative maximum,
and clipped to $[0,1]$. CarSim station is sampled at
$t=0,0.1,\ldots,8.0\,$s, shifted to start at zero, and divided by its maximum
over the same support. The same processing applies to the sunny control. The
weather-response error is
\begin{equation}
E_{\mathrm{weather}}(c)=\operatorname{RMSE}\!\left[
(\widehat p_c-\widehat p_0),
(p_c^{\mathrm{CS}}-p_0^{\mathrm{CS}})\right].
\end{equation}
Condition errors are averaged within each seed before the reported mean and
sample standard deviation ($\mathrm{ddof}=1$) are computed across seeds.
Because each monocular trajectory is independently normalized, this metric
evaluates the temporal profile of normalized braking progress rather than
absolute traveled distance. It is reported as a video-based response diagnostic
against the condition-specific normalized CarSim reference.

For muSync-GS, a replay check compares every seeded output with the
reference physical rollout. The branch-wise audit has 30 runs: five rain-branch
conditions (including sunny) and five snow-branch conditions (including sunny),
each over three seeds. The visual-quality set contains 27 unique videos because
the two branch-wise sunny controls are deduplicated. Across the 30 audit runs, the maximum
absolute difference is zero for all 19 logged physical and pose-driving
channels, including precipitation/depth controls, effective friction, speed,
station, pitch, heave, slip, road excitation, brake pressure, and source-frame
position. Thus, the reported variation isolates visual stochasticity rather
than rerunning or perturbing the vehicle response.

The pose-matched particle residual is computed from paired renders decoded in
8-bit display-color space. It is added to the pose-corrected static render and
clipped to the valid $[0,255]$ intensity range before encoding.

\paragraph{Artifact and seed release.}
Upon publication, we will release the implementation, preprocessing and
evaluation scripts, frozen configurations, run manifests, and all random seeds
under a license permitting free research use. Each stochastic result will be
paired with its configuration and seed, including seeds 2025--2027 for visual
generation and seed 20260725 for surface sensitivity; deterministic runs will
be marked explicitly.

\paragraph{Computing infrastructure.}
The Linux-side reconstruction, Gaussian rendering, and video evaluation were
run on a workstation with two NVIDIA RTX PRO 6000 Blackwell Max-Q Workstation
Edition GPUs (96 GB each), an Intel Xeon w9-3575X CPU (44 physical cores and
88 threads), and 128 GB of system RAM under Ubuntu 24.04.2 LTS. The Gaussian
stack used Python 3.10.19 and PyTorch 2.9.1 with CUDA 12.8; the VGGT and VACE
environments used Python 3.10.20 and PyTorch 2.8.0 with CUDA 12.8. Reference
vehicle-dynamics traces were generated with CarSim 2022.1. The planned artifact
release will include environment files and manifests recording task-specific
dependencies and model checkpoints.

\paragraph{CarSim-pose video reference.}
For each geometry-response task, the CarSim-pose reference uses the same road
profile, feature location, and source route as the corresponding video task.
The video speed-bump reference therefore uses the specified
$0.60\,\mathrm{m}\times0.10\,\mathrm{m}$ profile.
For each geometry case, we render the same reconstructed scene and road asset
with the original pose-neutral camera route, then apply the exact matched
CarSim speed, station, body-pitch, and CG-height channel $Z_{\mathrm{cg}}$. The reported pitch
channel is total CarSim body pitch, while the pose update applies only
$\theta_{\mathrm{CS}}(t)-\theta_0$ with $\theta_0=0.154150^\circ$; the static
orientation is therefore not applied twice.
Dynamic heave is
$Z_{\mathrm{cg}}-\operatorname{median}(Z_{\mathrm{cg}}[t<0.8\,\mathrm{s}])$.
The full exports contain 10,001 samples at $0.001\,$s intervals; before
rendering, their speed, station, and pitch are required to match the frozen
CarSim traces underlying main-paper Table~3 within $10^{-7}$, and the observed maximum differences are
zero for both cases. We render 81 frames at 10 fps and apply the same VGGT
camera-only estimator, fixed global pitch sign, event window, and baseline
flanks used for every geometry result, without temporal shift or sign/amplitude
fitting. The rendered clips are then encoded with the same H.264,
$480\times320$, 81-frame, 10-fps settings used by the muSync-GS videos evaluated
in main-paper Table~3.
The resulting video-bump/depression NRMSEs are $0.354/0.406$. They differ from
muSync-GS by at most $0.010$, providing an empirical rendering and
monocular-recovery reference rather than a strict mathematical lower bound.
This CarSim-pose video reference is distinct from the CarSim state-space reference
in the ablation, which bypasses rendering and VGGT.

For geometry response, the physical event window is fixed to
$2.2$--$3.8\,\mathrm{s}$. A local
linear baseline is estimated from the fixed flanking intervals
$[1.70,2.15]$ and $[3.85,4.30]\,\mathrm{s}$. All VGGT pitch curves use the
fixed second-order Savitzky--Golay smoothing specified by the evaluation
protocol before the local baseline fit. The CarSim reference is sampled on the
corresponding fixed time grid and is neither smoothed nor shifted. The
VGGT camera-pitch sign is converted once to the
CarSim vehicle-pitch convention. The normalized error is
\begin{equation}
E_{\mathrm{geometry}}=\frac{\operatorname{RMSE}
(\widehat\theta,\theta^{\mathrm{CS}})}
{\operatorname{RMS}(\theta^{\mathrm{CS}})},
\end{equation}
so a zero predicted event response gives $E_{\mathrm{geometry}}=1$. No
method-specific temporal shift, sign fitting, or amplitude fitting is
permitted. For each motion-preservation weather, the matching braking pitch is subtracted from
the video-bump/depression pitch separately for the input and output; each difference
is linearly detrended using frames outside 16--42. Source-motion retention is
the absolute input--output Pearson correlation on frames 16--42
($1.6$--$4.2\,$s at 10 fps), averaged over the six road-event clips within each
seed and then across seeds. Higher correlation indicates stronger retention of
motion already supplied in the motion-preservation input; it is a preservation
diagnostic rather than a counterfactual-generation score. It is not reported for
muSync-GS because the method does not take the motion-preservation video as an editor
input, so no like-for-like input--output pair exists.
\begin{table*}[t]
\centering
\small
\setlength{\tabcolsep}{3.3pt}
\begin{tabular}{p{0.12\textwidth}p{0.18\textwidth}p{0.21\textwidth}p{0.19\textwidth}p{0.21\textwidth}}
\toprule
Method & Weather input & Geometry input & Preservation input & Frozen inference settings \\
\midrule
Cosmos-Transfer2.5-2B & Sunny RGB, source depth, and a condition reference image &
Flat RGB, source depth, local depth-application mask, and flat/feature reference &
Condition-specific RGB and depth; context frame 28 & Official depth checkpoint;
35 steps, guidance 3;
depth weights 0.75 for weather response, 0.70 for geometry response, and
0.90 for motion preservation \\
LTX-Video 13B 0.9.8 & Sunny RGB and condition reference; no depth or mask &
Flat RGB and flat/feature reference at frame 28; no depth or mask &
Condition-specific RGB only & Fixed prompt without enhancement; 30 steps;
initial-step skips $17/17/24$, respectively \\
Wan2.1-VACE-14B & Depth-conditioned sunny video and condition reference &
Flat RGB, local edit mask, and flat/feature reference &
Condition-specific depth video and reference & 50 UniPC steps, guidance 5,
shift 16; plain prompt mode \\
\bottomrule
\end{tabular}
\vspace{4pt}

\parbox{0.96\textwidth}{\small
\emph{Frozen artifact identifiers.} Model revisions are
\url{nvidia/Cosmos-Transfer2.5-2B@dea7737ca29dd8d9086413c6dc5724b8250a0bb4}
(depth file \url{general/depth/626e6618-bfcd-4d9a-a077-1409e2ce353f_ema_bf16.pt}),
\url{Lightricks/LTX-Video@8984fa25007f376c1a299016d0957a37a2f797bb}
(13B 0.9.8 dev plus its 0.9.8 spatial upscaler), and
\url{Wan-AI/Wan2.1-VACE-14B@539c162b1387eac9dc4c20bd3f74671309e76a4c}.
The archived inference snapshots use \url{examples/inference.py},
\url{inference.py}, and \url{vace/vace_wan_inference.py}, and are based on
commits \texttt{2ff49d0}, \texttt{4b2d053}, and
\texttt{48eb44f}, respectively. The released Cosmos and LTX snapshots also
include the Blackwell depth-attention fallback and multi-scale video-to-video
media-resize compatibility patch used by these runs, so their upstream commits
alone do not fully specify execution.}
\caption{Native conditioning and frozen generation settings for the weather
response, geometry response, and motion-preservation video protocols.}
\label{tab:baseline_conditioning_supp}
\end{table*}

\subsection{Visual-control metrics}
\label{app:visual_protocol}

Weather appearance is compared at the same source scene, video time, and seed;
it is not used as evidence of correct vehicle response. Road-geometry metrics
use each geometry-response feature video and its paired flat control. Frames are first
registered by background-only ECC affine motion, after which a paired residual
mask is extracted inside a broad road region:

\begin{itemize}
    \item \textbf{Target-localized edit rate} is the fraction of target-visible
    frames with a detectable paired change in the prescribed road region. It
    measures localization, not semantic realism.
    \item \textbf{Road-anchor error} is the distance between detected and
    prescribed mask centroids, normalized by image diagonal. Missing detections
    receive an error of one.
    \item \textbf{Temporal mask IoU (T-IoU)} warps the preceding edit mask to
    the next frame using optical flow. Missing masks count as zero.
\end{itemize}

The binary target mask uses grayscale threshold 127 and is active at
$\geq80$ pixels. ECC affine registration excludes an 18-pixel target dilation
and the lower 12\% of the frame, with 70 iterations, tolerance $10^{-5}$, and
Gaussian filter size 5. The paired residual is mean absolute Lab difference,
thresholded at
$\max[0.025,\operatorname{median}_{\mathrm{bg}}+3.5\max(\operatorname{MAD}_{\mathrm{bg}},0.004)]$
inside the target-expanded lower-central road search region. A $3\times3$
opening and $7\times5$ closing are applied; a detection requires a connected
component of at least 20 pixels overlapping the expanded target and at least
5\% target coverage. T-IoU uses Farneback backward flow with pyramid scale
0.5, four levels, window 21, four iterations, $\texttt{poly\_n}=7$, and
$\texttt{poly\_sigma}=1.5$. Metrics are averaged over active frames within a
clip, then over the two road tasks within each seed; reported generator
uncertainty is the three-seed sample SD ($\mathrm{ddof}=1$).

Background LPIPS is excluded from the primary road table because global
weather transfer and the intended physics-aware viewpoint response can
legitimately change pixels outside a local road mask. For camera-path
preservation, each VGGT trajectory is independently normalized by its forward
extent and interpolated with its sunny counterpart on a common
normalized-progress grid $u\in[0,1]$. If $\widetilde y_c(u)$ and
$\widetilde y_0(u)$ are their normalized lateral coordinates, the clip error is
\begin{equation}
E_{\mathrm{path}}(c)=
\operatorname{RMS}_{u\in[0,1]}
\left[\widetilde y_c(u)-\widetilde y_0(u)\right].
\end{equation}
This prevents different braking durations from being counted as lateral drift.
For every method, the eight non-sunny conditions are macro-averaged within each
seed, followed by the mean and sample SD across seeds.

\section{Complete Vehicle-State Update}
\label{app:complete_vehicle_update}

\subsection{Reference surface coefficient}

The CarSim comparisons use the precipitation-derived coefficient
$\mu_c$ defined in the main paper. Flat-road weather and brake cases start at
$60\,\mathrm{km/h}$, road-profile cases at $40\,\mathrm{km/h}$, and the
separate road-profile speed check at $35/45\,\mathrm{km/h}$. The frozen
$\eta_{\mathrm{tire}}$ and $\beta_\kappa$ values are selected from $\mu_c$ and
held fixed through each stop.

\subsection{Wheel, brake, and ABS update}

For axle $a\in\{f,r\}$, the kinematic braking slip and horizontally scaled
tire-curve coordinate are
\begin{equation}
\kappa_a=\operatorname{clip}\!\left(
\frac{v-R\omega_a}{\max(v,v_\epsilon)},0,1\right),
\qquad \widetilde\kappa_a=\beta_{\kappa,w}\kappa_a,
\end{equation}
where $v_\epsilon=0.5\,\mathrm{m/s}$ is the slip denominator floor and
$\beta_{\kappa,w}$ is interpolated within the weather family. The longitudinal
force is
\begin{equation}
F_{x,a}=\mu_{\mathrm B}(\widetilde\kappa_a;\mu_{\mathrm{eff}})F_{z,a}.
\end{equation}
Let
\begin{equation}
\begin{aligned}
b(\kappa)&=c_1(1-e^{-c_2\kappa})-c_3\kappa,\\
\mu_{\mathrm B}(\kappa;\mu_{\mathrm{eff}})
&=\frac{\mu_{\mathrm{eff}}}{\max_{\lambda\in[0,1]}b(\lambda)}b(\kappa).
\end{aligned}
\end{equation}
This is the peak-normalized Burckhardt curve~\cite{burckhardt1993}. The
$(c_1,c_2,c_3)$ tuples are $(1.2801,23.99,0.52)$ for dry asphalt,
$(0.857,33.822,0.347)$ for wet asphalt, and
$(0.1946,94.129,0.0646)$ for snow. Scaling changes the peak to
$\mu_{\mathrm{eff}}$ without changing the base peak-slip location. Its target slip is
$\kappa_a^\star=\arg\max_{\kappa}
\mu_{\mathrm B}(\beta_{\kappa,w}\kappa;\mu_{\mathrm{eff}})$.
The pressure command is a scenario-defined ramp/hold/release profile. It is
mapped to axle torque by
\begin{equation}
T_b^{\mathrm{cmd}}=K_p p_b(t),\qquad
T_{b,a}=b_aq_aT_b^{\mathrm{cmd}},
\end{equation}
with $K_p=1000\,\mathrm{N\,m/MPa}$, $b_f=0.65$, and $b_r=0.35$. The formal
controller uses the proportional ABS command
\begin{equation}
q_a^{\mathrm{cmd}}=
\operatorname{clip}\!\left(
\frac{\kappa_a^\star(1+\delta_a)-\kappa_a}
{\delta_a\kappa_a^\star},0,1\right),
\label{eq:app_abs_command}
\end{equation}
where $\kappa_a^\star$ is the frozen target slip and $\delta_a=0.2$. The
reported configuration uses algebraic pressure response,
$q_a=q_a^{\mathrm{cmd}}$; the implementation also exposes the more general
rate-limited update
\begin{equation}
\begin{aligned}
\Delta q_a^n&=\operatorname{clip}\!\left(q_a^{\mathrm{cmd}}-q_a^n,
-r_{\mathrm{rel}}\Delta t,r_{\mathrm{build}}\Delta t\right),\\
q_a^{n+1}&=\operatorname{clip}\!\left(q_a^n+\Delta q_a^n,0,1\right),
\end{aligned}
\end{equation}
whose build and release rates are set sufficiently large to recover the
algebraic rule in all reported runs. Wheel rotation, longitudinal speed, and
path position obey
\begin{equation}
\begin{aligned}
I_{w,a}\dot\omega_a&=RF_{x,a}-T_{b,a},\\
m\dot v&=-(F_{x,f}+F_{x,r})+F_{x,\mathrm{road}}
-ma_{\mathrm{coast}},\\
F_{x,\mathrm{road}}&=-F_{z,f}H'_g(s)
-F_{z,r}H'_g(s-L_w),\\
\dot s&=v.
\end{aligned}
\label{eq:app_longitudinal_complete}
\end{equation}
At the stopping threshold $v\leq0.05\,\mathrm{m/s}$, the rollout sets
$v=\omega_f=\omega_r=\kappa_f=\kappa_r=0$ and resets the ABS torque scales.

\subsection{Four-DOF half-car and contact forces}

The half-car coordinates are
$\mathbf{x}_h=(z_s,\theta,z_{u,f},z_{u,r})$. Define
\begin{equation}
\begin{aligned}
d_f&=z_s-z_{u,f}-l_f\theta,
&F_{s,f}&=k_{s,f}d_f+c_{s,f}\dot d_f,\\
d_r&=z_s-z_{u,r}+l_r\theta,
&F_{s,r}&=k_{s,r}d_r+c_{s,r}\dot d_r.
\end{aligned}
\end{equation}
Because the coordinates are perturbations about static equilibrium, the
static axle loads are
\begin{equation}
F_{z,f}^0=m_sg\frac{l_r}{L_w}+m_{u,f}g,\qquad
F_{z,r}^0=m_sg\frac{l_f}{L_w}+m_{u,r}g.
\end{equation}
The tire-force increments and actual unilateral contact loads are
\begin{equation}
\begin{aligned}
F_{t,a}&=\max\!\left(k_{t,a}(z_{r,a}-z_{u,a}),-F_{z,a}^0\right),\\
F_{z,a}&=F_{z,a}^0+F_{t,a}.
\end{aligned}
\label{eq:app_unilateral_contact}
\end{equation}
The complete four-equation model is
\begin{equation}
\begin{aligned}
m_s\ddot z_s&=-F_{s,f}-F_{s,r},\\
I_y\ddot\theta&=l_fF_{s,f}-l_rF_{s,r}+M_b,\\
m_{u,f}\ddot z_{u,f}&=F_{s,f}+F_{t,f},\\
m_{u,r}\ddot z_{u,r}&=F_{s,r}+F_{t,r},
\end{aligned}
\label{eq:app_half_car_complete}
\end{equation}
with braking moment
\begin{equation}
M_b=\gamma_{\theta,w}m_sh_{\mathrm{cg}}d_\theta(t).
\label{eq:app_brake_pitch}
\end{equation}
Away from the stopping tail, $d_\theta(t)=\max(-\dot v,0)$. The half-car is
integrated by RK4 using linearly interpolated front and rear road inputs.

\subsection{Low-speed pitch rule and coupled update order}

The low-speed rule modifies only the brake-pitch input in
Eq.~\ref{eq:app_brake_pitch}; it does not modify simulated speed or stopping
distance. Let $t_s$ be the first sample with $v\leq0.05\,\mathrm{m/s}$ and
$t_h=\max(t_b,t_s-\tau_h)$, where $t_b$ is brake onset and $\tau_h$ is the
frozen family-dependent hold duration. We compute a reference deceleration
$\bar d_h$ as the median positive deceleration over the preceding
$0.6\,\mathrm{s}$. If the median over the final $0.1\,\mathrm{s}$ before
$t_s$ is below $\rho_h\bar d_h$, then $d_\theta(t)=\bar d_h$ on
$[t_h,t_s)$; otherwise the final $0.2\,\mathrm{s}$ is capped at
$1.05\bar d_h$. After stopping, the held value decays linearly to zero over
the frozen release duration $\tau_r$. The reported configuration uses
$\rho_h=0.8$; all $\tau_h$ and $\tau_r$ values are listed in
Table~\ref{tab:frozen_empirical_params}.

Road-feature cases use the same half-car equations and parameter set for road
and braking inputs. The implementation evaluates
\begin{equation}
\mathbf{x}_h=\mathbf{x}_{h,\mathrm{road}}+
\mathbf{x}_{h,\mathrm{brake}}.
\end{equation}
All evaluated cases remain in the active-contact regime, so the unilateral
clamp in Eq.~\ref{eq:app_unilateral_contact} is inactive and linear superposition applies to the common half-car system. The combined state supplies both axle loads and rendered camera
heave/pitch. For road-feature cases, the solver performs five global
trajectory-level fixed-point passes over the full internal time grid. Each pass:
\begin{enumerate}
    \item evaluate $z_{r,f/r}$ from the current path position;
    \item integrate the common half-car and compute $F_{z,f/r}$;
    \item update tire force, wheel rotation, speed, and path position using
    Eqs.~\ref{eq:app_abs_command}--\ref{eq:app_longitudinal_complete}.
\end{enumerate}
Flat-road braking uses acceleration-dependent longitudinal load transfer in
the wheel solver,
\begin{equation}
\begin{aligned}
F_{z,f}^{\mathrm{long}}
&=\max\!\left(m\frac{gl_r-\dot v h_{\mathrm{cg}}}{L_w},0\right),\\
F_{z,r}^{\mathrm{long}}
&=\max\!\left(m\frac{gl_f+\dot v h_{\mathrm{cg}}}{L_w},0\right).
\end{aligned}
\end{equation}
The common half-car separately provides the transient normal-load, heave, and
pitch signals used for evaluation and rendering. Thus dynamic half-car loads
feed the tire solver in road-feature cases, whereas flat-road longitudinal tire
forces use the analytic load-transfer equation above.
Accordingly, ``normal-load feedback'' denotes the case-appropriate load path:
dynamic half-car loads for road-feature cases and analytic longitudinal load
transfer for flat-road braking. It does not imply that transient half-car loads
are inserted into the flat-road tire solver.
All mechanical values, calibrated scalars, and numerical step sizes are given
in Tables~\ref{tab:frozen_physical_params} and
\ref{tab:frozen_empirical_params}.

\section{CarSim Development and Frozen Controlled-Holdout Results}
\label{app:carsim_ood}

\subsection{Calibration procedure and freezing}

We distinguish specified mechanical values from calibration parameters.
Mass, axle geometry, suspension coefficients, wheel properties, and the common
effective tire stiffness are taken from the matched CarSim vehicle setup or its
effective reduced-order representation (Table~\ref{tab:frozen_physical_params}).
The continuous calibration comprises 40 stored scalar entries, counted in
Table~\ref{tab:frozen_calibration_count}. The eight $\mu_c$ knot locations are
fixed by the development surface conditions and are not counted as calibrated
variables. Discrete structural choices---the common half-car/contact model,
local contact-slope road force, and zero road-type-specific energy
correction---were selected by development-stage structural ablations and then
frozen; they are reported separately and excluded from the scalar count.

Calibration proceeded iteratively by response group using the original-time
diagnostic vector
\begin{equation}
\begin{split}
\mathbf e_{\mathrm{cal}}=(&\operatorname{RMSE}_v,
\operatorname{RMSE}_\theta,\operatorname{RMSE}_\kappa,\\
&\operatorname{RMSE}_{F_z},|\Delta d_{\mathrm{stop}}|,
|\Delta v_{\mathrm{center}}|),
\end{split}
\end{equation}
as the common evaluation record. Each parameter group was adjusted against its directly affected response as
listed in Table~\ref{tab:frozen_calibration_count}. Normal-load RMSE was monitored
throughout as a cross-check.
The three structural switches were inspected in development-stage structural
comparisons before the empirical entries were fixed; this section reports the
selected structure rather than an additional quantitative ablation result.
Slip stiffness was adjusted
using per-wheel slip RMSE; longitudinal efficiency used flat-road speed and
stopping-distance errors; and brake-pitch gains and low-speed hold/release
parameters used body-pitch and stop-tail responses. The shared coupled
pitch scale was subsequently checked on the road-feature cases. After each
accepted update, all 19 development cases were rerun and the speed, slip,
pitch, normal-load, and event-level diagnostics were examined jointly. The
final entries and their associated diagnostic responses are reported in
Tables~\ref{tab:frozen_calibration_count} and
\ref{tab:frozen_empirical_params}.

The 19 development cases comprise nine flat-road braking cases at
$60\,\mathrm{km/h}$---dry, rain $2/10/25/50\,\mathrm{mm/h}$, and snow
$0.5/1/2/3\,\mathrm{mm/h}$---plus ten road-feature cases at
$40\,\mathrm{km/h}$. The latter apply five surface conditions (dry, rain 10,
rain 50, snow 1, and snow 3) to each of a
$75\,\mathrm{mm}\times3.7\,\mathrm{m}$ speed hump and an
$80\,\mathrm{mm}\times2.0\,\mathrm{m}$ road depression. All use the specified
$2.3\,\mathrm{MPa}$ brake command and $0.5\,\mathrm{s}$ pressure ramp for their
respective protocol.

The four initial-speed cases in Table~\ref{tab:carsim_development_supp} are held
out from calibration. The later 12-case controlled holdout suite uses the same frozen
configuration, with the four speed holdouts excluded from parameter selection.
No learned or additive time-series residual correction is applied in the
reported development or holdout results.

\begin{table}[!t]
\centering
\small
\setlength{\tabcolsep}{2.0pt}
\begin{tabular}{p{0.30\columnwidth}cp{0.46\columnwidth}}
\toprule
Calibration group & Scalars & Response used during staged selection \\
\midrule
Tire-force efficiency $\eta_{\mathrm{tire}}$ & 8 & Original-time speed and
stopping-distance errors in flat braking \\
Slip-stiffness scale $\beta_\kappa$ & 8 & Per-wheel slip-ratio RMSE \\
Brake-pitch scale $\gamma_\theta$ & 8 & Body-pitch RMSE during braking \\
Low-speed hold/release and hold ratio & $4+8+1$ & Stop-tail and release
portion of the body-pitch response \\
Coupled brake-pitch scale & 1 & Body-pitch RMSE in the ten road-feature
development cases \\
Static total-pitch offset & 1 & Initial total-pitch level read from the matched
CarSim development traces \\
Coast-drag acceleration $a_{\mathrm{coast}}$ & 1 & Shared development-frozen
coasting response \\
\midrule
Total & 40 & All entries frozen before held-out evaluation \\
\bottomrule
\end{tabular}
\caption{Accounting of the 40 scalar calibration entries. The third column
records the diagnostic response used during staged selection.}
\label{tab:frozen_calibration_count}
\end{table}

\begin{table*}[t]
\centering
\small
\setlength{\tabcolsep}{3.5pt}
\begin{tabular}{p{0.20\textwidth}p{0.18\textwidth}p{0.18\textwidth}p{0.34\textwidth}}
\toprule
Component & Parameter & Value & Role/provenance \\
\midrule
Mass and geometry & $m_s$; $m_{u,f},m_{u,r}$; $l_f,l_r$; CG height &
$1200\,\mathrm{kg}$; $54/54\,\mathrm{kg}$; $1.2/1.5\,\mathrm{m}$;
$0.55\,\mathrm{m}$ & Matched CarSim vehicle; total modeled mass is
$1308\,\mathrm{kg}$ and wheelbase is $2.7\,\mathrm{m}$. \\
Sprung-body rotation & $I_y$ & $1800\,\mathrm{kg\,m^2}$ & Shared pitch
inertia for braking and road excitation. \\
Suspension & $k_{s,f},k_{s,r}$; $c_{s,f},c_{s,r}$ &
$27{,}000/27{,}000\,\mathrm{N/m}$;
$3920/3920\,\mathrm{N\,s/m}$ & CarSim per-side values of
$13.5\,\mathrm{N/mm}$ and $1960\,\mathrm{N\,s/m}$ combined per axle. \\
Tire contact & $k_{t,f},k_{t,r}$; contact constraint &
{\footnotesize $193{,}268.620855\,\mathrm{N/m}$; unilateral} & Shared effective axle
tire stiffness; negative tire loads are disallowed. \\
Wheel & $R$; rotational inertia & $0.32\,\mathrm{m}$;
$2.4\,\mathrm{kg\,m^2}$ per axle & Effective rolling and rotational
parameters. \\
Brake/ABS & Pressure--torque gain; front bias; release width &
$1000\,\mathrm{N\,m/MPa}$ total; $0.65$; $0.20$ & The common development
command is $2.3\,\mathrm{MPa}$ with a $0.5\,\mathrm{s}$ ramp; held-out commands are
listed separately. \\
Road coupling & Longitudinal model; $a_{\mathrm{coast}}$; road-energy gain &
Local normal-load--slope force; $0.004\,\mathrm{m/s^2}$; $0$ & Local
front/rear contact slopes; the shared coast term is counted in
Table~\ref{tab:frozen_calibration_count}; no road-type-specific energy subtraction. \\
Integration/\allowbreak evaluation & Internal/output step; slip speed floor; duration &
$0.0005/0.001\,\mathrm{s}$; $0.5\,\mathrm{m/s}$; $10\,\mathrm{s}$ & Fixed
across matched runs. \\
\bottomrule
\end{tabular}
\caption{Fixed mechanical, controller, solver, and shared road-coupling
settings in the final frozen configuration. Per-axle suspension values combine
the two sides of the matched CarSim vehicle setup.}
\label{tab:frozen_physical_params}
\end{table*}

\begin{table*}[t]
\centering
\small
\setlength{\tabcolsep}{5pt}
\begin{tabular}{@{}lrrrrrr@{}}
\toprule
Family & $\mu_c$ & $\eta_{\mathrm{tire}}$ & $\beta_\kappa$ &
$\gamma_\theta$ & $\tau_h$ (s) & $\tau_r$ (s) \\
\midrule
Dry  & 0.820000 & 1.0100 & 1.65 & 1.104778 & 0.20 & 0.08 \\
\midrule
Rain & 0.347126 & 0.9278 & 1.00 & 1.133213 & 0.20 & 0.06 \\
     & 0.369279 & 0.9320 & 1.00 & 1.129315 & 0.20 & 0.06 \\
     & 0.434150 & 0.9427 & 1.00 & 1.123202 & 0.20 & 0.06 \\
     & 0.595585 & 1.0000 & 1.63 & 1.105687 & 0.20 & 0.08 \\
\midrule
Snow & 0.240000 & 0.9588 & 1.00 & 1.142519 & 0.20 & 0.06 \\
     & 0.420000 & 0.9932 & 1.00 & 1.111734 & 0.20 & 0.04 \\
     & 0.510000 & 0.9835 & 1.00 & 1.107972 & 0.05 & 0.06 \\
\bottomrule
\end{tabular}

\vspace{2pt}
\parbox{0.94\textwidth}{\small
Additional frozen scalars are static total pitch $0.154150^\circ$, low-speed
hold ratio $0.8$, and coupled brake-pitch scale $0.98$. These terms reproduce
the reported vehicle-response configuration. The displayed dry/rain
$\tau_h=0.20\,\mathrm{s}$ values share one stored scalar; together with three
snow-specific values, the hold-time group therefore contains four rather than
eight independent entries.}
\caption{Frozen calibration parameters. Within each weather family, the arrays
are listed one row per ascending $\mu_c$ knot and are interpolated
piecewise linearly; values outside the knot range use the nearest endpoint.
$\eta_{\mathrm{tire}}$ scales the surface coefficient, $\beta_\kappa$ scales
slip stiffness, $\gamma_\theta$ scales brake-pitch moment, and
$\tau_h/\tau_r$ are low-speed hold/release durations. The knot locations are fixed and
are not part of the 40-parameter count.}
\label{tab:frozen_empirical_params}
\end{table*}

\begin{table*}[t]
\centering
\small
\setlength{\tabcolsep}{4.5pt}
\begin{tabular}{p{0.16\textwidth}p{0.29\textwidth}p{0.27\textwidth}p{0.19\textwidth}}
\toprule
Control & Development values & Held-out values & Relationship \\
\midrule
Weather severity & Rain $2/10/25/50$; snow $0.5/1/2/3\,\mathrm{mm/h}$ &
Rain $5/15/35$; snow $0.75/1.5\,\mathrm{mm/h}$ & Unseen within-family
interpolation \\
Peak brake pressure & $2.3\,\mathrm{MPa}$ & $1.6\,\mathrm{MPa}$ & Lower
off-grid command \\
Pressure ramp & $0.5\,\mathrm{s}$ & $0.2\,\mathrm{s}$ & Faster off-grid
command \\
Brake onset & Time trigger at $1.0\,\mathrm{s}$ & Route-position trigger at
$12\,\mathrm{m}$ & Unseen trigger type \\
Speed hump & $75\,\mathrm{mm}\times3.7\,\mathrm{m}$ &
$50\,\mathrm{mm}\times3.7\,\mathrm{m}$;
$75\,\mathrm{mm}\times2.5\,\mathrm{m}$ & Off-grid height or length \\
Road depression & $80\,\mathrm{mm}\times2.0\,\mathrm{m}$ &
$50\,\mathrm{mm}\times2.0\,\mathrm{m}$;
$80\,\mathrm{mm}\times3.0\,\mathrm{m}$ & Off-grid depth or length \\
Initial speed (separate check) & $40\,\mathrm{km/h}$ for road cases &
$35/45\,\mathrm{km/h}$ & Local bidirectional off-grid check; not in the
12-case suite \\
\bottomrule
\end{tabular}

\vspace{5pt}
\begin{minipage}[t]{0.57\textwidth}
\centering
\small
\emph{Original development and initial-speed-check agreement (mean
event-window RMSE).}\\[2pt]
\setlength{\tabcolsep}{2.4pt}
\resizebox{\linewidth}{!}{%
\begin{tabular}{@{}lcccc@{}}
\toprule
Set & Pitch ($^\circ$) & Speed (m/s) & Slip & \shortstack{Per-wheel\\$F_z$ (N)} \\
\midrule
Flat braking (9)     & 0.0475 & 0.0288 & 0.00958 & 16.77 \\
Speed hump (5)       & 0.0354 & 0.0171 & 0.00150 & 25.16 \\
Road depression (5) & 0.0335 & 0.0185 & 0.00127 & 41.65 \\
All development (19)& 0.0406 & 0.0230 & 0.00527 & 25.53 \\
Speed holdout (4)    & 0.0263 & 0.0113 & 0.00011 & 22.35 \\
\bottomrule
\end{tabular}
}
\end{minipage}
\hfill
\begin{minipage}[t]{0.37\textwidth}
\small
For the nine development braking cases, braking-distance, braking-time, and
slip-distance MAEs are $0.107\,\mathrm{m}$, $0.0065\,\mathrm{s}$, and
$0.0535\,\mathrm{m}$, respectively. The four initial-speed cases use the same
frozen calibration and are not included in the 12-case suite.
\end{minipage}
\caption{Development--evaluation control relationship and original
development agreement. ``Interpolation'' is used only when held-out values lie
between observed development levels within the same weather family. A singleton
development setting cannot define an interpolation range, so the brake,
geometry, and initial-speed changes are reported as off-grid tests.}

\label{tab:carsim_development_supp}
\end{table*}

\subsection{Controlled holdout case manifest}

The 12-case suite uses the calibration configuration listed above and changes
three control families while retaining the reference vehicle and tire.

\begin{table*}[t]
\centering
\small
\setlength{\tabcolsep}{5pt}
\begin{tabular}{p{0.16\textwidth}p{0.24\textwidth}p{0.47\textwidth}}
\toprule
Held-out factor & Cases & Change relative to development protocol \\
\midrule
Weather severity & Rain $5/15/35\,\mathrm{mm/h}$; snow
$0.75/1.5\,\mathrm{mm/h}$ & New within-family levels. The frozen mapping gives
$\mu=0.503/0.401/0.356$ for rain and $0.465/0.330$ for snow; the same value is
passed to CarSim and muSync-GS. \\
Brake command & Peak pressure $1.6\,\mathrm{MPa}$; ramp $0.2\,\mathrm{s}$;
brake onset at $12\,\mathrm{m}$ along the route & Changes magnitude, application rate, or
start trigger while retaining the same surface, vehicle, tire, and calibration
parameters. \\
Road-elevation profile & Humps $50\,\mathrm{mm}\times3.7\,\mathrm{m}$ and
$75\,\mathrm{mm}\times2.5\,\mathrm{m}$; depressions
$50\,\mathrm{mm}\times2.0\,\mathrm{m}$ and
$80\,\mathrm{mm}\times3.0\,\mathrm{m}$ & Changes feature height or depth,
length, and slope through the same signed raised-cosine interface. \\
\bottomrule
\end{tabular}
\caption{Twelve-case controlled holdout design. All calibration parameters remain fixed during
evaluation. The weather cases use held-out precipitation levels under the
common frozen precipitation-to-surface-coefficient mapping. Snow rates are
water equivalent with the one-hour accumulation setting.}
\label{tab:carsim_ood_cases_supp}
\end{table*}

The weather-severity holdouts evaluate frozen reduced-order vehicle response
at unseen within-family precipitation levels under shared surface coefficients.
Appendix~\ref{app:surface_sensitivity} separately reports bounded sensitivity
to the surface-mapping parameters.

\subsection{Per-case controlled-holdout agreement}

\begin{table*}[t]
\centering
\small
\setlength{\tabcolsep}{4.0pt}
\begin{tabular}{llrrrrr}
\toprule
Split & Case & Speed (m/s) & Pitch ($^\circ$) & Slip & \shortstack{Per-wheel\\$F_z$ (N)} & Event error \\
\midrule
Weather & Rain 5             & 0.09862 & 0.10208 & 0.036701 & 31.53 & 0.28361 m \\
Weather & Rain 15            & 0.02910 & 0.05128 & 0.019291 & 15.26 & 0.14348 m \\
Weather & Rain 35            & 0.02420 & 0.04858 & 0.018267 & 14.35 & 0.11429 m \\
Weather & Snow 0.75          & 0.05342 & 0.08417 & 0.013246 & 37.00 & 0.14826 m \\
Weather & Snow 1.5           & 0.01649 & 0.04342 & 0.016212 & 13.42 & 0.01591 m \\
\midrule
Brake & Peak pressure 1.6 MPa & 0.01629 & 0.07403 & 0.005623 & 26.72 & 0.02099 m \\
Brake & Ramp 0.2 s            & 0.02406 & 0.09470 & 0.005945 & 36.04 & 0.02028 m \\
Brake & Onset at 12 m along route & 0.02372 & 0.05307 & 0.005553 & 21.99 & 0.01353 m \\
\midrule
Road profile & Hump $50\times3700$ mm       & 0.00979 & 0.03031 & 0.000040 & 22.10 & 0.03694 km/h \\
Road profile & Hump $75\times2500$ mm       & 0.01111 & 0.04135 & 0.000106 & 37.68 & 0.04028 km/h \\
Road profile & Depression $50\times2000$ mm & 0.01068 & 0.03143 & 0.000062 & 28.71 & 0.03878 km/h \\
Road profile & Depression $80\times3000$ mm & 0.01042 & 0.05318 & 0.000071 & 34.57 & 0.04058 km/h \\
\bottomrule
\end{tabular}
\caption{Per-case frozen controlled-holdout agreement. State columns are RMSE. The
last column is stopping-distance absolute error for weather and brake cases
and feature-center speed absolute error for road-geometry cases.}
\label{tab:carsim_ood_per_case_supp}
\end{table*}

\begin{figure}[!t]
\centering
\includegraphics[width=0.50\columnwidth]
{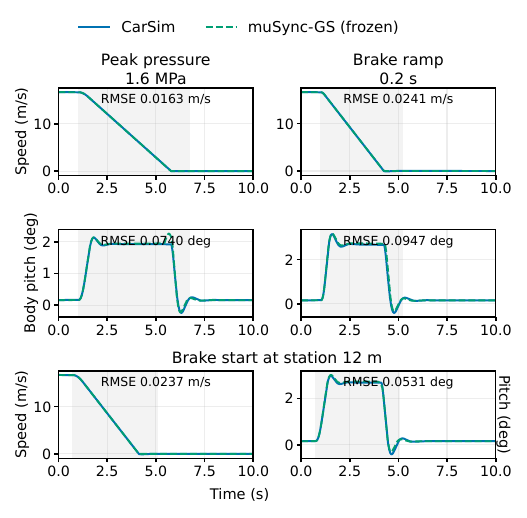}
\caption{All three frozen brake-input holdouts. The $0.2\,\mathrm{s}$ ramp has
the largest pitch transient error in this split, while stopping-distance error
remains below $0.021\,\mathrm{m}$ in every case.}
\label{fig:carsim_ood_brake_supp}
\end{figure}

\begin{figure}[!t]
\centering
\includegraphics[width=0.49\columnwidth]
{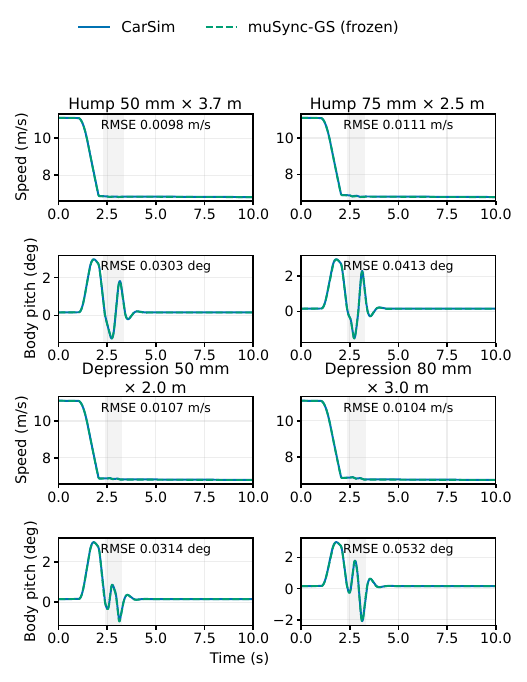}
\caption{All four frozen road-elevation-profile holdouts. New feature heights,
depths, lengths, and slopes are evaluated with the calibration frozen.}
\label{fig:carsim_ood_geometry_supp}
\end{figure}

The road-geometry split gives the strongest state transfer. The largest errors are
short weather- and brake-induced pitch transients: Rain 5, Snow 0.75, peak
pressure $1.6\,\mathrm{MPa}$, and ramp $0.2\,\mathrm{s}$ reach
$0.0740$--$0.1021^\circ$ pitch RMSE. These cases remain in every aggregate;
their smaller speed and event errors indicate that the main discrepancy is
transient brake dive or release rather than accumulated stopping behavior.
Complete trajectories following the evaluation rules in
Appendix~\ref{app:carsim_protocol} are shown in
Figures~\ref{fig:carsim_ood_weather_supp}--
\ref{fig:carsim_ood_geometry_supp}.

\begin{figure}[!t]
\centering
\includegraphics[width=0.78\columnwidth]
{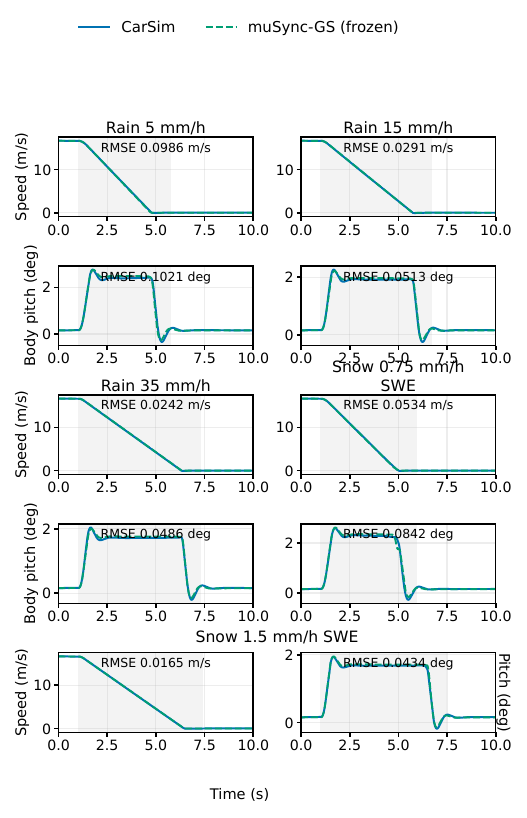}
\caption{All five frozen weather-severity holdouts under shared surface
coefficients derived from held-out precipitation levels. Gray shading denotes
the event window. Curves use original time and body pitch, including the static
offset, without alignment or calibration updates.}
\label{fig:carsim_ood_weather_supp}
\end{figure}

\begin{figure}[!t]
\centering
\includegraphics[width=0.70\columnwidth]
{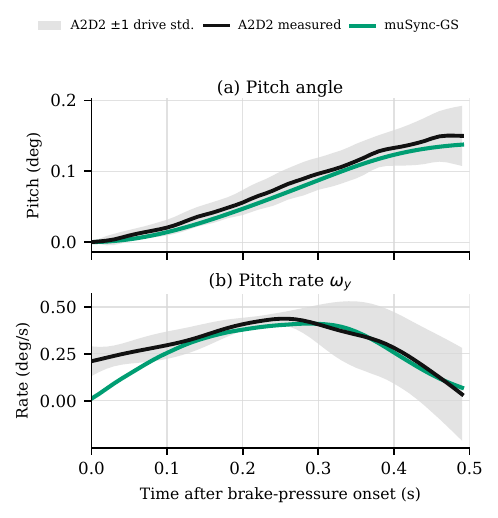}
\caption{Equal-drive macro A2D2 response for the joint pitch-angle-and-rate
calibration used in the main paper. Curves average the three held-out drive
means, and shading is $\pm1$ standard deviation across those drive means.
Calibration uses only the other two drives in each fold; held-out inference
uses brake pressure and event-start speed. All axes remain in physical units.}
\label{fig:a2d2_lodo_supp}
\end{figure}

\subsection{Cross-drive A2D2 brake-dive check after vehicle-specific calibration}
\label{app:a2d2_real_braking}

The A2D2 vehicle bus provides synchronized brake pressure, vehicle speed,
pitch angle, pitch rate, and acceleration~\cite{geyer2020a2d2}. Pressure
episodes are defined by samples at or above $1.2\,\mathrm{bar}$; consecutive
active samples separated by at most $0.08\,$s are merged. Of 140 detected
episodes in three Audi Q7 drives, 138 support the full $[-0.8,1.0]\,$s analysis
window and 14 satisfy all criteria listed below. The three drives contribute 8, 2, and
4 retained events, with event-start speeds from $13.1$ to
$36.4\,\mathrm{km/h}$. Pitch is expressed as nose-down change relative to a
linear pre-event baseline, and the evaluation window covers the first
$0.5\,$s after brake-pressure onset.

Brake onset is the first sample with pressure at or above $1.2\,$bar. A common
$[-0.6,-0.1]\,$s pre-event window supplies the median pressure and longitudinal
acceleration baselines and the affine pitch baseline. Pressure increase is the
maximum nonnegative baseline-subtracted pressure over $[0,0.5]\,$s; deceleration
increase is the maximum negative longitudinal-acceleration change over the same
window. All 12 selection criteria are applied conjunctively; no model prediction
or pitch-agreement score enters event selection.

We use complete-drive leave-one-drive-out evaluation. For each fold, a shared
set of effective vehicle parameters and the pressure gain are estimated using
only the other two drives. The angle-only ablation fixes the effective damping
ratio at $\zeta=0.70$. The joint pitch-angle-and-rate calibration used for the
main-paper result additionally estimates an effective damping ratio from the
two training drives in each fold. The half-car, pressure-to-slip, tire, and
controller equations are unchanged.
Held-out inference receives the measured
brake-pressure trace and event-start speed only. Neither measured pitch nor
$\omega_y$ is supplied at test time. The main paper reports equal-drive
macro statistics, while Figure~\ref{fig:a2d2_lodo_supp} shows the corresponding
equal-drive macro trajectory for the joint calibration. Fold sizes and the
angle-only calibration ablation are reported in the surrounding text.

The main-paper visualization shows the mean pitch trajectory of each held-out
drive. Within each panel, the measured and predicted drive means are
independently normalized by their positive peaks solely for temporal-shape
visualization; $q_\theta$ is computed from the corresponding unnormalized
peaks. Event selection, calibration, and aggregate metrics still use all 14
events, and all numerical metrics---including the macro curves in
Figure~\ref{fig:a2d2_lodo_supp}---are reported in physical units.

Here, \emph{calibration signals} are the measured signals from the two training
drives used to estimate the fold-specific parameters; they are not provided as
inputs for the held-out drive. ``Pitch only'' uses pitch angle as the calibration
signal, whereas ``Pitch + rate'' jointly uses pitch angle and pitch rate. The
main-paper result corresponds to the joint pitch-angle-and-rate calibration
with damping estimated only from the two training drives in each fold.

The fold-specific pressure-to-wheel-torque gain is estimated on the two
training drives by minimizing the residual between pressure-driven predicted
deceleration and baseline-corrected measured longitudinal deceleration. This
fit does not use pitch angle or pitch rate. The effective pitch-response
parameters are calibrated separately: the pitch-only variant uses training-drive
pitch angle, whereas the joint variant uses both pitch angle and pitch rate.
During this calibration, measured training-drive deceleration is used as the
half-car excitation to isolate the pitch dynamics.

The main paper reports that joint pitch-angle-and-rate
calibration attains drive-macro pitch RMSE
$0.062^\circ$ and correlation $0.901$, with pitch-rate RMSE
$0.381^\circ/\mathrm{s}$ and correlation $0.560$. These results support the direction and early temporal evolution of brake dive
after calibration on the other two drives of the same vehicle.

\begin{table}[H]
\centering
{\small
\textbf{A2D2 event-selection rules (138 complete-window candidates)}\\[2pt]
\setlength{\tabcolsep}{2.4pt}
\begin{tabular}{p{0.48\columnwidth}p{0.25\columnwidth}r}
\toprule
Criterion & Threshold & Fails \\
\midrule
Active-pressure duration & $\geq0.5\,\mathrm{s}$ & 45 \\
Start speed & $\geq12\,\mathrm{km/h}$ & 47 \\
Pressure increase & $\geq3.0\,\mathrm{bar}$ & 44 \\
Deceleration increase within $0.5\,$s & $\geq0.4\,\mathrm{m/s^2}$ & 41 \\
Absolute steering p95 & $\leq10^\circ$ & 71 \\
Absolute yaw-rate p95 & $\leq1.5^\circ/\mathrm{s}$ & 48 \\
Absolute pre-event pitch slope & $\leq0.25^\circ/\mathrm{s}$ & 64 \\
Pre-event pitch SD & $\leq0.07^\circ$ & 11 \\
Onset pitch residual & $\leq0.12^\circ$ & 45 \\
Abs. vertical-acceleration p95 & $\leq0.8\,\mathrm{m/s^2}$ & 8 \\
Absolute roll-rate p95 & $\leq1.5^\circ/\mathrm{s}$ & 40 \\
Pre-event pressure p95 & $\leq0.9\,\mathrm{bar}$ & 51 \\
\bottomrule
\end{tabular}\\[2pt]
\parbox{0.92\columnwidth}{\small Failure counts overlap; p95 denotes the 95th
percentile over $[-0.2,0.6]\,$s around pressure onset.}
}
\end{table}

\FloatBarrier

\section{Additional Video and Visual Results}
\label{app:video_visual_results}

\subsection{Video-observed motion curves}

\begin{figure}[!t]
\centering
\includegraphics[width=\columnwidth]
{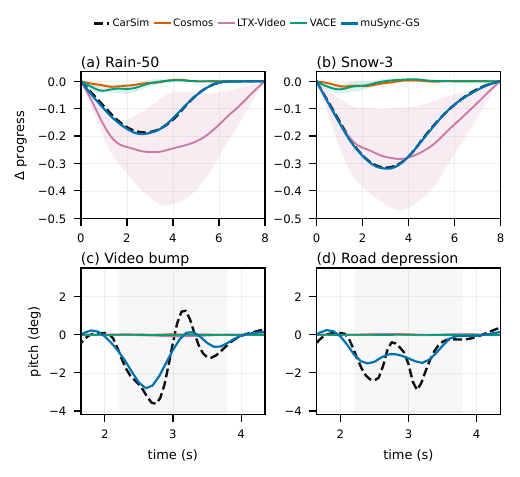}
\caption{VGGT-recovered motion against CarSim. (a)--(b) Weather-minus-sunny
normalized progress for Rain-50 and Snow-3. All baseline curves are means over
seeds 2025--2027, with bands showing one standard deviation. (c)--(d) Event-local
pitch for the video speed bump and road depression. Gray shading marks the fixed
$2.2$--$3.8\,\mathrm{s}$ event window.}
\label{fig:vggt_carsim_motion_supp}
\end{figure}

Figure~\ref{fig:vggt_carsim_motion_supp} visualizes the CarSim-referenced
weather- and geometry-response measurements summarized numerically in the main
paper. The heavy-weather
panels plot weather-minus-sunny normalized progress. The geometry panels plot
event-local pitch after the fixed baseline and sign conversion defined in
Appendix~\ref{app:video_protocol}. Baseline curves show the mean and variation over all three seeds. Although the
weather-response prompts do not request preservation of the source trajectory,
the baseline outputs remain close to the sunny-source motion on average, with
LTX showing the largest seed-dependent variation. For geometry, the baseline
prompts specify the feature location and dimensions but provide no target
camera motion; the near-zero recovered curves indicate that the visible
geometry edits are generally not accompanied by a corresponding pitch event.
In contrast, muSync-GS follows the localized CarSim response shape.

\paragraph{Five-level weather grid.}
The complete sunny, light-rain, heavy-rain, light-snow, and heavy-snow
comparison is already shown in the main-paper
Fig.~\ref{fig:weather_controllability} and is therefore not repeated here.

\subsection{Multi-scene qualitative consistency}

\begin{figure*}[!t]
\centering
\includegraphics[width=\textwidth]
{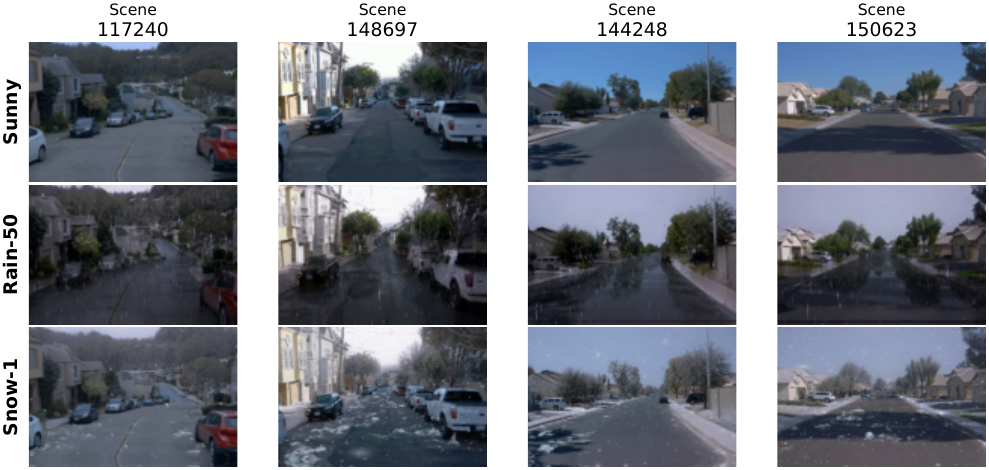}
\caption{Weather effects across four distinct reconstructed Waymo scenes from
one frozen rendering release. Rows show Sunny, Rain-50, and Snow-1; columns show
the scene IDs. All images use frame 28. The weather controls are shared, while
visible road support follows each scene's reconstruction and road mask.}
\label{fig:multiscene_weather_supp}
\end{figure*}

The main quantitative video protocols use Waymo context
9385013624094020582 over 2547.650--2567.650 s, denoted scene 938501.
The four segments below are additional qualitative scenes rendered with the
same frozen release and sampled at frame 28.
Figure~\ref{fig:multiscene_weather_supp} extends the weather visualization to
four distinct reconstructed Waymo segments~\cite{sun2020waymo}. The selected scenes contain only static or slowly moving surrounding agents,
with no nearby interaction-critical actor whose trajectory would require
counterfactual resimulation. The columns correspond to scene IDs
117240, 148697, 144248, and 150623, rather than multiple viewpoints from
one route. Each column uses the same scene-specific rendering assets across the
three weather conditions. The labels denote common nominal controls; visible
road support remains scene dependent because each reconstruction has its own
road mask, texture, geometry, and viewing configuration.

\subsection{Automatic quality and severity metrics}
\label{app:automatic_visual_metrics}

The primary automatic evaluation contains 128 videos. The weather subset has
108 videos: each of the four methods uses nine conditions and three visual
seeds. The road subset has 20 videos: two tasks and three seeds per baseline
(18), plus two deterministic muSync-GS outputs. For muSync-GS weather, only
visual stochastic variables change across seeds; the physical state and
camera trajectory are frozen. DOVER, FAST-VQA, CLIP severity, and temporal
LPIPS use the 108-video weather subset; localization, anchor, and T-IoU use the
20-video road subset. For normalized no-reference and temporal visual-quality
evaluation, videos are encoded as H.264 at $480\times320$, 10 fps, and 81
frames. The first 81 frames of each native 16-fps geometry output are preserved
in order and retimed, so these clip-normalization diagnostics are not
physical-time response measures. The road-mask, anchor, and T-IoU diagnostics
remain frame-indexed on the same ordered 81 source frames. As stated in
Appendix~\ref{app:baseline_conditioning}, the VGGT geometry-response metric
instead uses the native 16-fps timestamps.

DOVER~\cite{wu2023dover} is the primary no-reference video-quality metric;
FAST-VQA-B~\cite{wu2022fastvqa} is a robustness check with overlapping scope.
Both use the authors' official pretrained checkpoints and score-fusion code.
For semantic weather severity, we uniformly sample 16 frames and encode them
with CLIP ViT-B/32~\cite{radford2021clip}. Four-prompt ensembles describe dry,
heavy-rain, and heavy-snow dashcam scenes. The exact dry prompts are ``a clear
dry road seen from a dashcam,'' ``a sunny driving scene with no precipitation,''
``a dry suburban road viewed through a windshield,'' and ``clear weather in an
on-board driving video.'' The rain prompts are ``a dashcam driving scene in
heavy rain,'' ``a road seen through a windshield during heavy rainfall,'' ``an
on-board driving video with intense rain,'' and ``a wet road in a severe
rainstorm.'' The snow prompts are ``a dashcam driving scene in heavy snow,'' ``a
road seen through a windshield during heavy snowfall,'' ``an on-board driving
video with intense snow,'' and ``a snow-covered road in a severe snowstorm.''
With normalized frame feature $f_t$
and mean text feature $g_w$, the rain score is
\begin{equation}
s_{\mathrm{rain}}=\frac{1}{16}\sum_{t=1}^{16}
\left(f_t^\top g_{\mathrm{rain}}-f_t^\top g_{\mathrm{dry}}\right),
\end{equation}
and snow is analogous. We report Spearman's $\rho$ between numeric input and
severity score over sunny plus four positive levels. Adjacent accuracy is the
fraction of consecutive pairs whose score increases. These are semantic
control metrics, not photorealism measures. We interpret them as a
severity-ordering sanity check rather than evidence of photorealism or
independent physical validity.

Table~\ref{tab:musync_weather_seed_supp} reports the individual muSync-GS
seed results. The low but nonzero variation in VGGT and no-reference quality
metrics is induced only by particles and spatial snow realization; CLIP
severity ordering is unchanged. VACE obtains the highest DOVER and FAST-VQA
values, while muSync-GS maintains monotonic semantic ordering across all tested
rain and snow control levels and obtains the lowest road-anchor error. Together,
these metrics distinguish perceptual
quality from continuous severity control and localized road consistency.

\begin{table}[!t]
\centering
\small
\setlength{\tabcolsep}{1.1pt}
\textbf{(a) Aggregate no-reference weather-video quality}\par\smallskip
\begin{tabular}{lcccc}
\toprule
Method & Tech. & Aesth. & Overall & FAST \\
\midrule
Cosmos & $-0.1361$ & $-0.0376$ & $0.1760$ & $0.4134$
\\
LTX-Video & $-0.1335$ & $-0.0233$ & $0.2033$ & $0.4014$
\\
VACE & $\mathbf{-0.1009}$ & $\mathbf{0.0345}$ & $\mathbf{0.3679}$
& $\mathbf{0.6514}$ \\
muSync-GS & $-0.1274$ & $-0.0476$ & $0.1800$ & $0.4545$
\\
\bottomrule
\end{tabular}

\vspace{6pt}
\textbf{(b) Severity ordering and road-anchor consistency}\par\smallskip
\begin{tabular}{lccccc}
\toprule
Method & Rain $\rho$ & Snow $\rho$ & Rain Adj. & Snow Adj. & Anchor $\downarrow$ \\
\midrule
Cosmos & $0.9667$ & $0.8667$ & $0.9167$ & $0.6667$ & $0.678$ \\
LTX-Video & $0.9333$ & $0.8667$ & $0.8333$ & $0.8333$ & $0.708$ \\
VACE & $0.9333$ & $0.6667$ & $0.8333$ & $0.7500$ & $0.440$ \\
muSync-GS & $\mathbf{1.0000}$ & $\mathbf{1.0000}$ & $\mathbf{1.0000}$
& $\mathbf{1.0000}$ & $\mathbf{0.296}$ \\
\bottomrule
\end{tabular}

\vspace{6pt}
\textbf{(c) Flow-aligned temporal LPIPS over positive weather levels}\par\smallskip
\begin{tabular}{lcccc}
\toprule
Method & Rain raw & Rain excess & Snow raw & Snow excess \\
\midrule
Cosmos & $0.0438$ & $0.0002$ & $0.0447$ & $0.0012$ \\
LTX-Video & $0.0385$ & $-0.0001$ & $0.0522$ & $0.0136$ \\
VACE & $0.0451$ & $0.0004$ & $0.0538$ & $0.0092$ \\
muSync-GS & $0.0734$ & $0.0510$ & $0.1482$ & $0.1258$ \\
\bottomrule
\end{tabular}
\caption{Automatic visual evaluation. Technical and Aesthetic are raw DOVER
fields; Overall is the fused DOVER score; FAST is FAST-VQA-B; Adj. is
adjacent-level ordering accuracy; Anchor is normalized road-anchor error.
Panels (a), the weather columns of (b), and (c) use the 108-video weather
subset; Anchor uses the 20-video road subset. The main paper reports three-seed
mean $\pm$ sample SD for its selected columns, and
Table~\ref{tab:musync_weather_seed_supp} gives the individual muSync-GS seeds.
The muSync-GS road output is deterministic. Flow-aligned roadside temporal
LPIPS is diagnostic because visible precipitation also increases the score.}
\label{tab:visual_quality_full_supp}
\end{table}

\subsection{Flow-aligned temporal LPIPS diagnostic}

We additionally sample 40 adjacent frame pairs per weather video. Farneback
backward flow warps the preceding frame to the current frame, and
LPIPS-Alex~\cite{zhang2018lpips} is evaluated in two fixed roadside regions,
$(0,60,150,240)$ and $(330,60,480,240)$. The matching method/seed sunny score
is subtracted to obtain excess temporal LPIPS.

This diagnostic is excluded from the main comparison because rain streaks and
snow particles legitimately cross static-scene regions. It therefore
combines intended moving precipitation with background flicker in a single
score.

\subsection{Interpretation of the visual metrics}

Together with the qualitative comparison, the road-geometry metrics
characterize edit localization, temporal persistence, and the associated
vehicle response. A high target-localized edit rate indicates that a change
appears in the intended region. Anchor error and T-IoU add spatial and temporal
evidence, while the geometry-response evaluation tests whether the visible
feature is accompanied by the expected vehicle response. The scene-preservation
metric is progress aligned so that differences in braking duration are not
counted as lateral camera drift.

\begin{table}[!t]
\centering
\small
\setlength{\tabcolsep}{2.5pt}
\textbf{(a) Recovered-motion errors}\par\smallskip
\begin{tabular}{lcc}
\toprule
Seed & \shortstack{Weather diagnostic\\RMSE} & \shortstack{Lateral RMSE\\$\downarrow$}
\\
\midrule
2025 & 0.004241 & 0.002600 \\
2026 & 0.004345 & 0.002732 \\
2027 & 0.004146 & 0.002653 \\
\midrule
Mean $\pm$ SD & \shortstack{0.004244\\$\pm0.000099$}
& \shortstack{0.002662\\$\pm0.000067$} \\
\bottomrule
\end{tabular}

\vspace{5pt}
\textbf{(b) No-reference quality}\par\smallskip
\begin{tabular}{lcc}
\toprule
Seed & DOVER Overall $\uparrow$ & FAST $\uparrow$ \\
\midrule
2025 & 0.1787 & 0.4542 \\
2026 & 0.1825 & 0.4547 \\
2027 & 0.1788 & 0.4546 \\
\midrule
Mean $\pm$ SD & $0.1800\!\pm\!0.0021$ & $0.4545\!\pm\!0.0003$ \\
\bottomrule
\end{tabular}

\vspace{5pt}
\textbf{(c) Severity rank correlation}\par\smallskip
\begin{tabular}{lcc}
\toprule
Seed & Rain $\rho$ & Snow $\rho$ \\
\midrule
2025 & 1.000 & 1.000 \\
2026 & 1.000 & 1.000 \\
2027 & 1.000 & 1.000 \\
\midrule
Mean $\pm$ SD & $1.000\!\pm\!0.000$ & $1.000\!\pm\!0.000$ \\
\bottomrule
\end{tabular}
\caption{Per-seed muSync-GS weather robustness. Weather RMSE is the VGGT
progress-response error; lateral RMSE is progress-aligned lateral RMS; DOVER
Overall and FAST are macro-averages over nine weather conditions. Only visual
stochastic variables change. Lateral RMSE first macro-averages the eight
non-sunny conditions within each seed, then reports the mean and sample standard
deviation across the three seeds.}
\label{tab:musync_weather_seed_supp}
\end{table}

\section{Bounded Surface-Parameter Sensitivity}
\label{app:surface_sensitivity}

\subsection{Scope of the analysis}

This analysis evaluates the reported weather--friction ordering across bounded
perturbations of the dense-asphalt parameter vector. Precipitation remains a
deterministic input, while the listed surface parameters are sampled over the
declared engineering ranges. Literature and standards provide the physical
scale or measurement basis for the selected nominal values. This is a bounded
model-sensitivity analysis rather than independent validation against measured
wet- or snow-road friction.

\subsection{Surface equations and parameter basis}

For rain, let $I_r$ be rainfall intensity in $\mathrm{mm/h}$, $\mathrm{MTD}$
be mean texture depth in $\mathrm{mm}$, $L_d$ drainage-path length in
$\mathrm{m}$, and $S$
crossfall in $\mathrm{m/m}$. Crossfall is substituted in decimal form when the
equation is evaluated; for example, $2.0\%$ corresponds to $S=0.020$. The metric
Gallaway relation gives
\begin{equation}
\begin{aligned}
 h_t &= 0.01485\,\mathrm{MTD}^{0.11}L_d^{0.43}I_r^{0.59}S^{-0.42},\\
 h_f &= \max(0,h_t-\mathrm{MTD}).
\end{aligned}
\end{equation}
Here $h_t$ is the total water depth, in millimeters, measured from the bottom of
the pavement macrotexture and used in the main-paper surface state. The derived
$h_f$, also in millimeters, is the nonnegative free-water depth above the
texture peaks and is retained only as a diagnostic. The reported rain-friction
mapping uses $h_t$ rather than treating $h_f$ as an additional fitted friction
input.
The variables and exponents originate from controlled pavement-drainage
experiments~\cite{gallaway1971waterdepth,gallaway1979hydroplaning}. The raw
rain-condition coefficient is
\begin{equation}
\begin{aligned}
 \mu_c^{\mathrm{rain}}&=\mu_{\mathrm{dry}}
 \left[r_f+(1-r_f)e^{-h_t/h_c}\right],\\
 r_f&=\mu_{\mathrm{flooded}}/\mu_{\mathrm{dry}},
\end{aligned}
\end{equation}
where $h_c$ is a film-decay scale in millimeters, applied to the same
bottom-referenced water-depth convention. For water-equivalent snowfall $I_s$
accumulated for duration $T$,
\begin{equation}
\begin{aligned}
 h_s&=I_sT\,\rho_w/\rho_s,\\
 c_s&=\min(1,h_s/h_{\mathrm{cover}}),\\
 \mu_c^{\mathrm{snow}}&=\mu_{\mathrm{bw}}-
 (\mu_{\mathrm{bw}}-\mu_{\mathrm{snow}})c_s.
\end{aligned}
\end{equation}
Here $\rho_w=1000\,\mathrm{kg/m^3}$ is the water density and $\rho_s$ is the
fresh-snow density; $c_s$ is a contact-coverage proxy, and
$h_{\mathrm{cover}}$ is a model threshold rather than a road-design standard.

\begin{table*}[!t]
\centering
\small
\setlength{\tabcolsep}{4pt}
\begin{tabular}{p{0.17\textwidth}p{0.12\textwidth}p{0.14\textwidth}p{0.49\textwidth}}
\toprule
Parameter & Nominal & Sampled bound & Source basis and interpretation \\
\midrule
Dry coefficient $\mu_{\mathrm{dry}}$ & $0.82$ & $[0.78,0.86]$ &
Frozen anchor with an approximately $\pm5\%$ perturbation. The nominal is
inside the $0.8$--$1.0$ dry-bare friction-number range summarized by
VTI~\cite[p.~36]{wallman2001roadfriction}. \\
Flooded coefficient $\mu_{\mathrm{flooded}}$ & $0.34$ & $[0.306,0.374]$ &
Frozen anchor with $\pm10\%$ perturbation. Wet-road measurements show a
broad surface-dependent range~\cite{ivan2010wetfriction}. \\
Texture depth $\mathrm{MTD}$ & $0.50\,\mathrm{mm}$ & $[0.40,0.60]\,\mathrm{mm}$ &
Dense-graded-asphalt nominal with $\pm20\%$ perturbation. Field MPD statistics
have a comparable scale~\cite{boz2023densegraded}. Volumetric MTD measurement
is standardized by ASTM E965/E965M~\cite{astmE965}. \\
Drainage length $L_d$ & $3.75\,\mathrm{m}$ & $[3.0,4.5]\,\mathrm{m}$ &
Scenario flow-path nominal with $\pm20\%$ perturbation; $L_d$ is the
route-dependent input to the Gallaway model~\cite{gallaway1971waterdepth}. \\
Crossfall $S$ & $2.0\%$ & $[1.6,2.4]\%$ &
$\pm20\%$ around the nominal. FHWA HEC-22 lists $1.5$--$2.0\%$ for high-type
two-lane surfaces and notes that $2.5\%$ may aid drainage in intense-rainfall
regions~\cite{kilgore2024hec22}. \\
Film-decay scale $h_c$ & $0.30\,\mathrm{mm}$ & $[0.24,0.36]\,\mathrm{mm}$ &
Model-specific transition scale with $\pm20\%$ perturbation. \\
Fresh-snow density $\rho_s$ & $100\,\mathrm{kg/m^3}$ & $[80,120]\,\mathrm{kg/m^3}$ &
$\pm20\%$ around the conventional $10{:}1$ water-to-snow nominal. Four-year
observations report site means of $72$--$103\,\mathrm{kg/m^3}$ and wider event
variability~\cite{judson2000freshsnowdensity}. \\
Full-cover threshold $h_{\mathrm{cover}}$ & $20\,\mathrm{mm}$ & $[16,24]\,\mathrm{mm}$ &
Model-specific tire-contact interpolation threshold with $\pm20\%$
perturbation. \\
Bare-wet transition $\mu_{\mathrm{bw}}$ & $0.60$ & $[0.54,0.66]$ &
Frozen transition anchor with $\pm10\%$ perturbation. It lies within broad
field wet-pavement measurements~\cite{ivan2010wetfriction}. \\
Snow coefficient $\mu_{\mathrm{snow}}$ & $0.24$ & $[0.216,0.264]$ &
$\pm10\%$ around the fresh-snow anchor, consistent with the $0.20$--$0.25$
new-snow skid-resistance range in controlled tests
~\cite{ichihara1970snowfriction}. \\
\bottomrule
\end{tabular}
\caption{Nominal surface parameters and declared sensitivity bounds.
Intervals are controlled perturbations unless explicitly stated otherwise.
MPD and volumetric MTD are related macrotexture measures but are not identical.}
\label{tab:surface_parameter_sources}
\end{table*}

Texture depth and crossfall support the dense-asphalt nominal, while drainage
length comes from the selected route. Film decay and full-cover depth are
explicit model transition scales.

\subsection{Sampling and frozen simulation protocol}

We generate 128 independent Latin-hypercube samples for each weather family
with seed 20260725. Rain samples
$\{\mu_{\mathrm{dry}},\mu_{\mathrm{flooded}},\mathrm{MTD},L_d,S,h_c\}$,
and snow samples
$\{\mu_{\mathrm{dry}},\mu_{\mathrm{bw}},\mu_{\mathrm{snow}},\rho_s,
h_{\mathrm{cover}}\}$, uniformly within the corresponding bounds in
Table~\ref{tab:surface_parameter_sources}. Rain uses 21 intensities from 0 to
$50\,\mathrm{mm/h}$; snow uses 13 water-equivalent rates from 0 to
$3\,\mathrm{mm/h}$ with $T=1\,\mathrm{h}$. Dense nominal curves use 51 and 31
rates, respectively. The experiment contains 4,434 solver cases.

All non-surface quantities are frozen to the common braking protocol: initial
speed $60\,\mathrm{km/h}$, total brake torque $2300\,\mathrm{N\,m}$, front
bias $0.65$, torque ramp $0.5\,\mathrm{s}$, ABS release width $0.2$, and
internal step $5\times10^{-4}\,\mathrm{s}$. Each raw coefficient passes
through the frozen family-specific tire efficiency and slip-stiffness
response terms before the same brake solver is run. Against archived nominal
outputs, the maximum raw-$\mu$ discrepancy is $4.3\times10^{-11}$ and the
maximum stopping-distance discrepancy is $1.37\,\mathrm{mm}$.

\begin{figure}[!t]
\centering
\includegraphics[width=\columnwidth]
{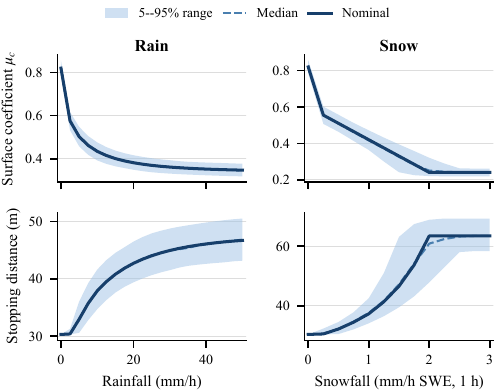}
\caption{Bounded surface-parameter sensitivity under the frozen
brake/ABS protocol. Top: raw condition coefficient $\mu_c$. Bottom: stopping
distance. Solid lines use nominal parameters; bands are the 5th--95th
percentiles over the 128 sampled parameter sets.}
\label{fig:surface_parameter_sensitivity_supp}
\end{figure}

\subsection{Envelope and rank-sensitivity results}

Across the evaluated positive snowfall rates, the sampled surface-coefficient
and stopping-distance curves remain monotonic. At Rain-50, nominal raw $\mu_c$ is $0.347$ with a
$0.318$--$0.377$ envelope, and stopping distance is $46.70\,\mathrm{m}$ with
a $43.16$--$50.50\,\mathrm{m}$ envelope. Stopping-distance envelopes are
$34.07$--$42.53\,\mathrm{m}$ at Snow-1 and
$58.39$--$69.35\,\mathrm{m}$ at saturated Snow-3.

Spearman correlations with stopping distance at Rain-50 are dominated by
$\mu_{\mathrm{flooded}}$ ($\rho=-0.973$). At Snow-1, the largest absolute
correlations are full-cover depth ($-0.601$), snow density ($-0.535$),
bare-wet friction ($-0.489$), and snow friction ($-0.354$). At Snow-3, full
coverage is saturated and the snow coefficient dominates ($\rho=-1.000$).
These rankings match the model equations: partial coverage depends on
accumulation and transition depth, whereas saturated coverage depends mainly
on the snow-contact friction anchor.

\FloatBarrier

\end{document}